\pdfoutput=1

\documentclass[11pt]{article}

\usepackage{acl}

\usepackage{times}
\usepackage{latexsym}
\usepackage{graphicx}
\usepackage{tabularx}
\usepackage{multirow}
\usepackage{amssymb}
\usepackage{amsmath}
\usepackage{xcolor,colortbl}
\usepackage{subcaption}
\usepackage{stfloats} 
\usepackage{hhline}
\usepackage[inline]{enumitem}
\usepackage{changepage}
\usepackage{adjustbox}
\usepackage{array}
\usepackage{rotating}

\usepackage{pgfplots}
\usepgfplotslibrary{colorbrewer}
\pgfplotsset{cycle list/Dark2}
\pgfplotsset{compat=1.18}

\newcolumntype{R}[2]{%
    >{\adjustbox{angle=#1,lap=\width-(#2)}\bgroup}%
    l%
    <{\egroup}%
}
\newcommand*\rot{\multicolumn{1}{R{40}{1em}}}
\newcommand\blfootnote[1]{%
  \begingroup
  \renewcommand\thefootnote{}\footnote{#1}%
  \addtocounter{footnote}{-1}%
  \endgroup
}

\newcommand{\camera}[1]{#1}
\newcommand{\rebuttal}[1]{#1}

\usepackage[T1]{fontenc}

\usepackage[utf8]{inputenc}

\usepackage{microtype}

\usepackage{booktabs}

\title{Morality is Non-Binary: Building a Pluralist Moral Sentence Embedding Space using Contrastive Learning}

\author{Jeongwoo Park$^\ast$, Enrico Liscio$^\ast$, and Pradeep K. Murukannaiah \\
Delft University of Technology, the Netherlands \\
\texttt{E.Liscio@tudelft.nl}}

\begin{document}
\maketitle

\blfootnote{$^\ast$ Equal Contribution.}

\begin{abstract}
Recent advances in NLP show that language models retain a discernible level of knowledge in deontological ethics and moral norms. However, existing works often treat morality as binary, \camera{ranging from right to wrong}. This simplistic view does not capture the nuances of moral judgment. Pluralist moral philosophers argue that human morality can be deconstructed into a finite number of elements, respecting individual differences in moral judgment. 
In line with this view, we build a pluralist moral sentence embedding space via a state-of-the-art contrastive learning approach. 
\rebuttal{We systematically investigate the embedding space by studying the emergence of relationships among moral elements, both quantitatively and qualitatively.}
Our results show that a pluralist approach to morality can be captured in an embedding space. 
\camera{However, moral pluralism is challenging to deduce via self-supervision alone and requires a supervised approach with human labels.}
\end{abstract}

\section{Introduction}
Morality helps humans distinguish right from wrong \cite{Graham2013}. As AI systems work with (or for) humans, it is crucial that they align with human morality \cite{Gabriel2020,Liscio2023AAMAS}. Several NLP methods have been proposed to recognize human morality in text \cite{forbes-etal-2020-social, Lourie_Le_Bras_Choi_2021, jiang2021delphi,pyatkin2023clarifydelphi}. However, such methods typically treat morality as a score that ranges in a single dimension of right to wrong. This does not reflect the nuances in moral reasoning, differences among individuals, or the existence of moral value conflicts \cite{implicationsofMF}.

Pluralist moral philosophers argue that morality should be represented through a finite number of basic elements, referred to as moral values \cite{Graham2013}. Each situation triggers one or more moral values, and each of us assigns varying importance to each moral value. The combination of these two aspects determines the individual moral judgment in the situation. For instance, the debate on immigration touches on the moral values of \textit{fairness} (``Everyone should be given equal opportunities'') and in-group \textit{loyalty} (``I worry about the preservation of our identity''). The way in which each of us prioritizes fairness vs. loyalty influences our moral judgment in this debate. Thus, morality cannot (and should not) be unidimensionally classified in text \cite{talat-etal-2022-machine}. Instead, the moral elements that are salient to a piece of text can be recognized, which can be used to reason about or assist the humans in the moral judgment.

The Moral Foundations Theory (MFT) is a popular pluralist approach to morality \cite{Graham2013} which states that people have five innate moral foundations on which they base their moral judgments.
There is a surge of interest \camera{in morality \cite{vida-etal-2023-values} and particularly in the} MFT in the NLP community \cite{kobbe-etal-2020-exploring, alshomary-etal-2022-moral, liscio-etal-2022-cross, Liscio2023}, partly due to the Moral Foundation Twitter Corpus (MFTC) \cite{Hoover2020}, composed of 35k tweets annotated with the MFT foundations.

Prior research has focused on methods for classifying MFT elements in a textual discourse \cite{Huang2022LearningWeighting,alshomary-etal-2022-moral,liscio-etal-2022-cross}. 
However, such methods provide limited qualitative insight into the relations between text and MFT elements.
We explore the mapping between text and MFT through sentence embeddings, which consist of a multi-dimensional representation that encapsulates knowledge from textual data. 
Instead of being limited to a specific task, a suitable sentence embedding space can be valuable across multiple NLP tasks, such as text classification, generation, and \rebuttal{topic modelling} \cite{henderson-etal-2020-convert,10.1145/3482853, zhang-etal-2022-neural}.
Further, a sentence embedding space can be geometrically explored, allowing us to investigate the relationships among different moral elements.

\citet{nmi} show that pre-trained sentence embeddings contain a moral direction that maps actions from ``do'' to ``don't'', without the need for re-training on morally loaded data. \rebuttal{In this work, we investigate whether the same holds for a pluralist approach to morality. That is: do pre-trained sentence embeddings contain discernible clusters corresponding to the different elements of a pluralist approach to morality, or is it necessary to re-train them with a supervised approach to disentangle the different moral elements?}

\rebuttal{Our contribution is twofold. First,} we propose a novel approach for mapping the MFT elements to a sentence embedding space using the state-of-the-art SimCSE \cite{gao-etal-2021-simcse} method, which makes use of the Contrastive Learning paradigm \cite{9226466}. 
\rebuttal{Then,} we evaluate the resulting embedding space in two ways. 
First, we perform an intrinsic evaluation to investigate the relationship between different moral elements and \rebuttal{evaluate whether a supervised approach is necessary to disentangle the MFT elements in the embedding space}. Second, \rebuttal{to evaluate whether the relationships among the MFT elements have been adequately captured}, we perform an extrinsic evaluation, \rebuttal{generalizing the analyses to a novel test set and to the set of words from a moral dictionary.}

Our experiments show that a pluralist approach to morality can be captured in a sentence embedding space, but also that human labels are necessary to successfully train the embeddings.
Our work represents the starting point for incorporating a pluralist approach to morality in language models, with a warning that self-supervision alone is not sufficient to capture the complexity of human morality.

\section{Background and Data}
\label{sec:background}
We introduce the method to train sentence embedding spaces (SimCSE) and the data we use.

\paragraph{SimCSE} 
Sentence embedding spaces represent sentences as points in a high-dimensional space,
mapping semantically similar sentences to the same region of space.
Contrastive Learning (CL) \cite{9226466} is an approach to training an embedding space based on a contrastive loss that aims to minimize the distance between positive (semantically similar) sentence pairs and maximize the distance between negative (semantically dissimilar) sentence pairs.
\rebuttal{Formally, let $x_i$ and $x_i^+$ be positively related and $\mathbf{h_i}$, $\mathbf{h_i^+}$ be their encoded representations. Then, the training loss for the two instances with a mini-batch of $N$ pairs is:}

\rebuttal{
\begin{equation}
\ell_i=-\log \frac{e^{\operatorname{sim}\left(\mathbf{h}_i, \mathbf{h}_i^{+}\right) / \tau}}{\sum_{j=1}^N e^{\operatorname{sim}\left(\mathbf{h}_i, \mathbf{h}_j^{+}\right) / \tau}}
\end{equation} where $\tau$ is a temperature hyperparameter and sim($\mathbf{h_1}$, $\mathbf{h_2}$) the cosine similarity \cite{gao-etal-2021-simcse}.}

SimCSE \cite{gao-etal-2021-simcse} is a text-based CL framework built on BERT sentence embeddings \cite{reimers-gurevych-2019-sentence} \rebuttal{that demonstrated better performance than other BERT variants \cite{gao-etal-2021-simcse}.}
SimCSE supports \emph{supervised} and \emph{unsupervised} approaches. 
Supervised SimCSE seeks to minimize the distance between sentences with the same label and maximize the distance between sentences with different labels. 
Unsupervised SimCSE generates a positive instance by applying a slight variation of a reference sentence through dropout, and uses a random sentence as a negative instance. 
\rebuttal{We detail the SimCSE supervised and unsupervised CL loss in Appendix~\ref{app:cl-loss}.}

\paragraph{Moral Foundations Twitter Corpus}
The MFT \cite{Graham2013} is a popular pluralist theory of morality that postulates that human morality is composed of five innate moral foundations
\rebuttal{that combine to describe our moral stance over divisive issues.}
Each of the five foundations of the MFT is composed of a virtue-vice duality, resulting in the 10 moral elements shown in Table~\ref{tab:mft}.

\begin{table}[!htb]
\centering
\small
 \begin{tabular}{@{}p{1.3cm}p{5.9cm}@{}}
 \toprule
 \textbf{Element} & \textbf{Definition} \\
 \midrule
 Care/\newline Harm & Support for care for others/\newline Refrain from harming others \\ 
 \midrule
 Fairness/\newline Cheating & Support for fairness and equality/\newline Refrain from cheating or exploiting others\\ 
 \midrule
 Loyalty/\newline Betrayal & Support for prioritizing one’s inner circle/\newline Refrain from betraying the inner circle\\
 \midrule
 Authority/\newline Subversion & Support for respecting authority and tradition/\newline Refrain from subverting authority or tradition \\
 \midrule
 Purity/\newline Degradation & Support for the purity of sacred entities/\newline Refrain from corrupting such entities \\
 \bottomrule
\end{tabular}
\caption{The MFT moral foundations (virtue/vice).}
\label{tab:mft}
\end{table}

The Moral Foundations Twitter Corpus (MFTC) \cite{Hoover2020} is a collection of 35,108 tweets collected in seven domains: All Lives Matter, Baltimore Protest, Black Lives Matter, hate speech and offensive language \cite{Davidson_2017}, 2016 presidential election, MeToo movement, and hurricane Sandy. The tweets were annotated with one or more of the 10 MFT elements, or with a \textit{non-moral} label. As each tweet was annotated by multiple annotators (ranging from 3 to 8), the authors of MFTC use a majority vote to choose the definitive label(s) of each tweet (thus resulting in one or more moral labels per tweet), and \textit{non-moral} is assigned when no majority is present.

\section{Training the Embedding Space}
\label{sec:method}
We train the moral embedding space by finetuning \textit{unsupervised} and \textit{supervised} SimCSE approaches. 
The unsupervised approach does not employ label information, thus the strategy described in Section~\ref{sec:background} is used.
In the supervised approach, SimCSE uses label information to construct the training triples for its supervised CL objective function.
Each triple is composed of 
\begin{enumerate*}[label=(\arabic*)]
    \item a \textit{reference} data point,
    \item a data point whose distance from the reference should be minimized (\textit{positive instance}), and
    \item a data point whose distance from the reference should be maximized (\textit{negative instance}).
\end{enumerate*}

\begin{figure}[!htb]
    \centering
    \includegraphics[width=0.75\columnwidth]{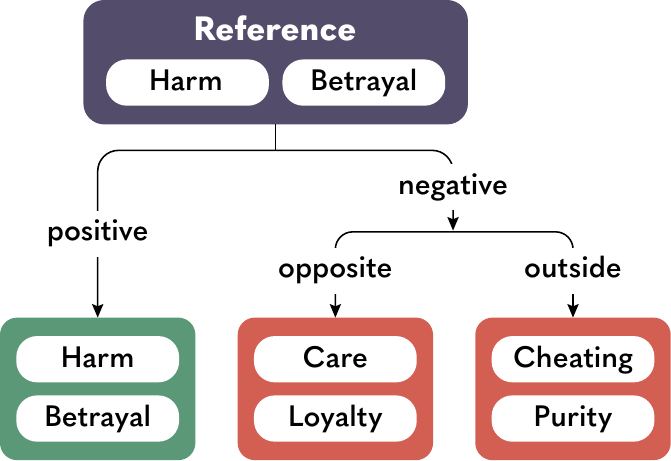}
    \caption{Example triple formation with the two policies for negative instance selection (\textit{opposite} and \textit{outside}).}
    \label{fig:sup_triple}
\end{figure}

Figure~\ref{fig:sup_triple} shows an example of how the triples are constructed. In this example, the chosen reference instance is labeled with two moral elements---\textit{harm} and \textit{betrayal}. Then, the positive instance is chosen as a data point with the same labels as the reference instance. However, selecting negative instances is not trivial due to the structure of the MFT taxonomy, which is composed of five pairs of virtue-vice. Thus, we propose two policies, \textit{opposite} and \textit{outside}, to guide the choice of negative instances.

The \textit{opposite} policy selects the negative instance as a data point annotated with moral elements that are opposite virtue/vice of the reference labels (\textit{care} and \textit{loyalty} in the example). In contrast, the \textit{outside} policy chooses the negative instance as a data point annotated with moral elements that belong to other moral foundations than the reference foundations (\textit{cheating} and \textit{purity} in the example).

In both policies, we prioritize data points with more negative labels when choosing the negative instance, when possible. For instance, in the example in Figure~\ref{fig:sup_triple}, with the \textit{opposite} policy, we prioritize a data point with the labels \textit{care} and \textit{loyalty} over a data point with just the \textit{care} label.
We divide the MFTC training set into two halves and apply each policy to a half. We ensure that each data point appears in just one triple.
When no suitable positive or negative instances are available, data points labeled as \textit{non-moral} are used as positive or negative instances, until all morally-loaded data points have been used in a triple.

\section{Evaluating the Embedding Space}
We use 90\% of the MFTC as the training set to train the moral embedding space \rebuttal{(with the approaches described in Section~\ref{sec:method})} and the remaining 10\% as the test set. To generate a balanced training (and test) set, we randomly selected 90\% (and 10\%) of data from each of the seven domains in MFTC, resulting in the label distribution in Table~\ref{tab:label_distr}.
Data pre-processing, hyperparameters, and training environment are detailed in Appendix~\ref{sec:appendix1}. \camera{The code is available on GitHub\footnote{\url{https://github.com/jeongwoopark0514/morality-is-non-binary}}}.

\begin{table*}[!htb]
\small
\centering
\begin{tabular}{@{}p{0.8cm}p{0.5cm}p{0.6cm}p{0.95cm}p{0.95cm}p{0.8cm}p{0.95cm}p{1.1cm}p{1.2cm}p{0.7cm}p{1.35cm}p{1.4cm}@{}}
\toprule
\textbf{Dataset} & Care & Harm & Fairness & Cheating & Loyalty & Betrayal & Authority & Subversion & Purity & Degradation & Non-moral \\ \midrule
\textbf{Train}      & 2176 & 3269 & 1870     & 3068     & 1736    & 1736     & 1294      & 1816       & 698    & 1246 & 14428        \\
\textbf{Test}       & 240  & 359  & 204      & 335      & 183     & 121      & 137       & 196        & 72     & 132  & 1611        \\ \bottomrule
\end{tabular}
\caption{Distribution of MFT labels in the training and test sets used to train and evaluate SimCSE moral embeddings.}
\label{tab:label_distr}
\end{table*}

\rebuttal{We first inspect the embedding space \camera{itself} to evaluate whether a supervised approach is needed to disentangle the MFT elements in the MFTC training set (intrinsic evaluation).
Then, to evaluate whether the relationships among MFT elements have been successfully captured, we test the embedding space on two \camera{downstream} tasks (as suggested by \citet{eger-etal-2019-pitfalls}) (extrinsic evaluation).}

\subsection{Intrinsic Evaluation}
\label{sec:experimental-setup-intrinsic}
\rebuttal{We investigate the embedding space by
\begin{enumerate*}[label=(\arabic*)]
    \item showing a visualization of the training set data in the embedding space to gain an intuitive understanding of the relationships among MFT elements, and
    \item computing a moral similarity table to inspect quantitative similarities among MFT elements.
\end{enumerate*}
To show the effect of supervised labels during training, we compare
\begin{enumerate*}[label=(\alph*)]
    \item an off-the-shelf pre-trained supervised SimCSE embedding space, and the embeddings trained with
    \item the unsupervised SimCSE and
    \item the supervised SimCSE approaches.
\end{enumerate*}}

\subsubsection{Visualization}
We explore the relationships between the MFT elements in the embedding space through visual insight. 
Since the SimCSE embedding space is 1024-dimensional, we employ the Uniform Manifold Approximation and Projection (UMAP) method \cite{mcinnes2020umap}, a nonlinear dimensionality reduction technique, to reduce the embedding space to two dimensions.
We choose UMAP as it preserves both local and most of the global structure in the data, with a shorter run-time when compared to other dimensionality reduction techniques such as t-SNE and PCA \cite{mcinnes2020umap}. 
We show all the data points in the MFTC training set in a two-dimensional plot
and qualitatively discuss the relationships among MFT elements.

\subsubsection{Moral Similarity}
\label{sec:moral-similarity-MFTC}

We perform a moral similarity task, inspired by the popular semantic similarity task \cite{agirre-etal-2013-sem, gao-etal-2021-simcse}, to measure the similarity between moral elements using the MFTC training set.
To calculate the moral similarity between two MFT elements $m$ and $n$, we compute the cosine similarity between the moral embedding representations of each data point annotated with $m$ and each data point annotated with $n$, and report the mean result. We apply the procedure for all combinations of the ten MFT elements plus the \textit{non-moral} label, resulting in an 11x11 table of mean similarities.

\subsection{Extrinsic Evaluation}
\label{sec:experimental-setup-extrinsic}

\rebuttal{To evaluate whether the relationships among MFT elements have been effectively captured in the embedding spaces, we evaluate \begin{enumerate*}[label=(\arabic*)]
    \item the generalizability on the held-out test set, and
    \item the consistency between the embeddings and the Moral Foundation Dictionary 2.0 (MFD2.0) \cite{Frimer_2019}, an independently collected MFT dictionary.
\end{enumerate*}
As in Section~\ref{sec:experimental-setup-intrinsic}, we compare
\begin{enumerate*}[label=(\alph*)]
    \item an off-the-shelf pre-trained SimCSE embedding space, and the embeddings trained with
    \item the unsupervised SimCSE and
    \item the supervised SimCSE approaches.
\end{enumerate*}}

\subsubsection{Generalizability on Test Set}
\rebuttal{We evaluate the moral embedding spaces on the MFTC test set to assess the generalizability to unseen data. As for the intrinsic evaluation described above, we evaluate the embedding spaces
\begin{enumerate*}[label=(\arabic*)]
    \item via a visualization by plotting the MFTC test set on the embedding space and visualizing it via a UMAP plot, and
    \item with a moral similarity table.
\end{enumerate*}}

\subsubsection{Comparison to MFD2.0}

We measure the consistency of the generated moral embedding spaces with MFD2.0, a dictionary manually created by the authors of the MFT \cite{Graham2013}, containing sets of words representative of each MFT moral element.

\paragraph{Clustering}
\label{sec:clustering}
We collect all words belonging to the MFD2.0 and use $K$-means clustering to test whether meaningful clusters can be discerned based on the words' embedding representations based on their Euclidean distance (we choose Euclidean since the $K$-means algorithm may not converge with other distances without data transformation).

First, we measure the coherence of the clusters via the silhouette coefficient \cite{rousseeuw1987silhouettes}:
\begin{equation}
    s = \frac{\sum_{i=1}^{N}\frac{b(i)-a(i)}{max(a(i),b(i))}}{N}
\end{equation}
where $N$ is the number of samples, $a(i)$ the mean intra-cluster distance and $b(i)$ the mean nearest-cluster distance for sample $i$. The coefficient ranges from -1 to 1.
For each tested approach, we plot the silhouette coefficient for $K$ ranging from 2 to 15 and choose $\hat{K}$ as the optimal number of clusters with the highest silhouette score.

Then, we measure the quality of the clusters via the purity score \cite{manning2009introduction}. To calculate the purity of a cluster, we first find the most frequent true label ($L_f$) of each cluster. Then, we sum the number of words labeled with $L_f$ for each cluster and divide the sum by the total number of words in the dictionary.
Thus, a high purity score indicates that the clusters primarily consist of words with the same label. However, the purity score tends to increase as $K$ increases, since each cluster is at the purest state when there is only one item in the cluster. 
Due to this tradeoff between $K$ and the clustering quality, we evaluate the clustering results via both the silhouette coefficient and the mean purity score over the clusters.
We report the results for $K=\hat{K}$ and $K=10$ (as the MFT taxonomy is composed of ten elements).

\paragraph{Moral Similarity \rebuttal{(MFD2.0)}}

We measure the similarity among the MFD2.0 words belonging to different MFT elements via moral similarity, as in Section~\ref{sec:moral-similarity-MFTC}.
To calculate the moral similarity between two MFT elements $m$
and $n$, we compute the cosine similarity between the moral embedding representations of each MFD2.0 word belonging to $m$ and each MFD2.0 word belonging to $n$, and report the mean result. We apply the procedure for all combinations of the ten MFT elements, resulting in a 10x10 table of mean similarity.

\section{Results and Discussion}
\label{sec:result}
We report the results of the intrinsic evaluations to judge the effect of supervised training, and the results of the extrinsic evaluation to assess the moral embeddings when used with external data.

\subsection{Intrinsic Evaluation}

We present the results of visualization and moral similarity evaluations on the MFTC training set.

\subsubsection{Visualization}
\label{sec:result-umap}
\begin{figure*}[!htb]
    \centering
    \includegraphics[width=\linewidth]{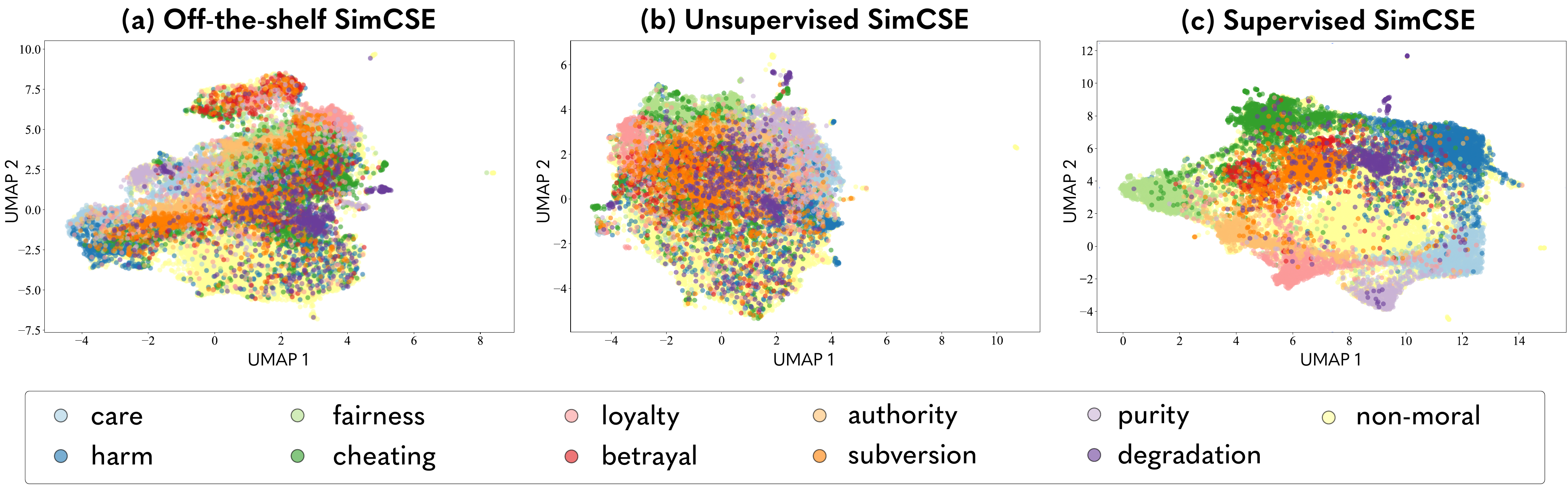}
    \caption{\rebuttal{UMAP plot of the \textbf{MFTC training set} data with off-the-shelf pre-trained SimCSE model (a, left), unsupervised SimCSE approach (b, middle), and supervised SimCSE approach (c, right).}}
    \label{fig:umap}
\end{figure*}

Figure~\ref{fig:umap} shows the dimension-reduced UMAP plot of the MFTC training set data mapped on the moral embedding spaces 
\rebuttal{
\begin{enumerate*}[label=(\alph*)]
    \item resulting from the off-the-shelf pre-trained supervised SimCSE model, or trained with
    \item the unsupervised SimCSE approach or
    \item the supervised SimCSE approach.
\end{enumerate*}}

We notice that the supervised approach (Figure~\ref{fig:umap}c) shows distinguishable clusters for each vice and virtue element, exhibiting a visible improvement when compared to the off-the-shelf model (Figure~\ref{fig:umap}a).
However, the unsupervised approach (Figure~\ref{fig:umap}b) displays no discernible clusters.

In Figure~\ref{fig:umap}c, we observe a clear separation between virtues (located in the bottom half of the plot) and vices (located in the top half). 
Further, the values within the same foundation (e.g., \textit{care-harm}) tend to be in symmetrical locations in the virtues and vices areas. 
Finally, tweets labeled as \textit{non-moral} are spread throughout the plot, especially in the area between the vice and virtue clusters. 

The noticeable difference between the off-the-shelf, unsupervised, and supervised approaches suggests that a CL-based moral embedding space can capture the relationships between virtues and vices and among moral foundations when employing label information.
\rebuttal{We investigate this further via a quantitative moral similarity evaluation.}

\subsubsection{Moral Similarity}
\label{sec:moral-similarity-mftc}

To further analyze the insightful results observed with the supervised approach, we report in Table~\ref{tab:moral-sim-MFTC} the moral similarity across MFT elements calculated with the supervised SimCSE moral embedding representations of the MFTC training set. This table allows us to inspect in more detail the similarity across the different moral elements.

\begin{table}[!htb]
\centering 
\hspace{-0.68cm}
\resizebox{0.98\linewidth}{!}{
\scriptsize
\renewcommand{\tabcolsep}{1.3pt}
\renewcommand{\arraystretch}{1.4} 
\begin{tabular}{rccccc|ccccc|c}
 & \rot{Care} & \rot{Fairness} & \rot{Loyalty} & \rot{Authority} & \rot{Purity} & \rot{Harm} & \rot{Cheating} & \rot{Betrayal} & \rot{Subversion} & \rot{Degradation} &
 \rot{Non-moral}\\
Care & {\cellcolor[HTML]{023858}} \color[HTML]{F1F1F1} 81.2 & {\cellcolor[HTML]{E3E0EE}} \color[HTML]{000000} 25.4 & {\cellcolor[HTML]{529BC7}} \color[HTML]{F1F1F1} 41.0 & {\cellcolor[HTML]{94B6D7}} \color[HTML]{000000} 35.2 & {\cellcolor[HTML]{0569A5}} \color[HTML]{F1F1F1} 49.5 & {\cellcolor[HTML]{D6D6E9}} \color[HTML]{000000} 27.6 & {\cellcolor[HTML]{FFF7FB}} \color[HTML]{000000} 4.7 & {\cellcolor[HTML]{F7F0F7}} \color[HTML]{000000} 21.0 & {\cellcolor[HTML]{FFF7FB}} \color[HTML]{000000} 15.2 & {\cellcolor[HTML]{FFF7FB}} \color[HTML]{000000} 11.6 & {\cellcolor[HTML]{CED0E6}} \color[HTML]{000000} 28.8 \\
Fairness & {\cellcolor[HTML]{E3E0EE}} \color[HTML]{000000} 25.4 & {\cellcolor[HTML]{023858}} \color[HTML]{F1F1F1} 77.9 & {\cellcolor[HTML]{CED0E6}} \color[HTML]{000000} 28.8 & {\cellcolor[HTML]{3991C1}} \color[HTML]{F1F1F1} 43.0 & {\cellcolor[HTML]{CCCFE5}} \color[HTML]{000000} 29.1 & {\cellcolor[HTML]{FFF7FB}} \color[HTML]{000000} 12.7 & {\cellcolor[HTML]{9AB8D8}} \color[HTML]{000000} 34.6 & {\cellcolor[HTML]{FEF6FB}} \color[HTML]{000000} 19.2 & {\cellcolor[HTML]{F1EBF5}} \color[HTML]{000000} 22.4 & {\cellcolor[HTML]{FFF7FB}} \color[HTML]{000000} 10.9 & {\cellcolor[HTML]{DDDBEC}} \color[HTML]{000000} 26.4 \\
Loyalty & {\cellcolor[HTML]{529BC7}} \color[HTML]{F1F1F1} 41.0 & {\cellcolor[HTML]{CED0E6}} \color[HTML]{000000} 28.8 & {\cellcolor[HTML]{023858}} \color[HTML]{F1F1F1} 65.0 & {\cellcolor[HTML]{7BACD1}} \color[HTML]{F1F1F1} 37.7 & {\cellcolor[HTML]{8BB2D4}} \color[HTML]{000000} 36.2 & {\cellcolor[HTML]{FFF7FB}} \color[HTML]{000000} 9.7 & {\cellcolor[HTML]{FFF7FB}} \color[HTML]{000000} 8.5 & {\cellcolor[HTML]{D6D6E9}} \color[HTML]{000000} 27.7 & {\cellcolor[HTML]{FEF6FB}} \color[HTML]{000000} 19.1 & {\cellcolor[HTML]{FFF7FB}} \color[HTML]{000000} 8.7 & {\cellcolor[HTML]{D9D8EA}} \color[HTML]{000000} 27.0 \\
Authority & {\cellcolor[HTML]{94B6D7}} \color[HTML]{000000} 35.2 & {\cellcolor[HTML]{3991C1}} \color[HTML]{F1F1F1} 43.0 & {\cellcolor[HTML]{7BACD1}} \color[HTML]{F1F1F1} 37.7 & {\cellcolor[HTML]{023858}} \color[HTML]{F1F1F1} 68.7 & {\cellcolor[HTML]{589EC8}} \color[HTML]{F1F1F1} 40.5 & {\cellcolor[HTML]{FFF7FB}} \color[HTML]{000000} 11.3 & {\cellcolor[HTML]{FFF7FB}} \color[HTML]{000000} 14.4 & {\cellcolor[HTML]{E3E0EE}} \color[HTML]{000000} 25.4 & {\cellcolor[HTML]{7EADD1}} \color[HTML]{F1F1F1} 37.4 & {\cellcolor[HTML]{FFF7FB}} \color[HTML]{000000} 14.1 & {\cellcolor[HTML]{D8D7E9}} \color[HTML]{000000} 27.3 \\
Purity & {\cellcolor[HTML]{0569A5}} \color[HTML]{F1F1F1} 49.5 & {\cellcolor[HTML]{CCCFE5}} \color[HTML]{000000} 29.1 & {\cellcolor[HTML]{8BB2D4}} \color[HTML]{000000} 36.2 & {\cellcolor[HTML]{589EC8}} \color[HTML]{F1F1F1} 40.5 & {\cellcolor[HTML]{023858}} \color[HTML]{F1F1F1} 79.3 & {\cellcolor[HTML]{FFF7FB}} \color[HTML]{000000} 13.2 & {\cellcolor[HTML]{FFF7FB}} \color[HTML]{000000} 5.2 & {\cellcolor[HTML]{FFF7FB}} \color[HTML]{000000} 15.5 & {\cellcolor[HTML]{FFF7FB}} \color[HTML]{000000} 17.5 & {\cellcolor[HTML]{F1EBF5}} \color[HTML]{000000} 22.4 & {\cellcolor[HTML]{D9D8EA}} \color[HTML]{000000} 27.2 \\
\hhline{~-----------}
Harm & {\cellcolor[HTML]{D6D6E9}} \color[HTML]{000000} 27.6 & {\cellcolor[HTML]{FFF7FB}} \color[HTML]{000000} 12.7 & {\cellcolor[HTML]{FFF7FB}} \color[HTML]{000000} 9.7 & {\cellcolor[HTML]{FFF7FB}} \color[HTML]{000000} 11.3 & {\cellcolor[HTML]{FFF7FB}} \color[HTML]{000000} 13.2 & {\cellcolor[HTML]{023D60}} \color[HTML]{F1F1F1} 56.9 & {\cellcolor[HTML]{D9D8EA}} \color[HTML]{000000} 27.2 & {\cellcolor[HTML]{91B5D6}} \color[HTML]{000000} 35.5 & {\cellcolor[HTML]{C2CBE2}} \color[HTML]{000000} 30.2 & {\cellcolor[HTML]{B5C4DF}} \color[HTML]{000000} 31.7 & {\cellcolor[HTML]{C4CBE3}} \color[HTML]{000000} 30.0 \\
Cheating & {\cellcolor[HTML]{FFF7FB}} \color[HTML]{000000} 4.7 & {\cellcolor[HTML]{9AB8D8}} \color[HTML]{000000} 34.6 & {\cellcolor[HTML]{FFF7FB}} \color[HTML]{000000} 8.5 & {\cellcolor[HTML]{FFF7FB}} \color[HTML]{000000} 14.4 & {\cellcolor[HTML]{FFF7FB}} \color[HTML]{000000} 5.2 & {\cellcolor[HTML]{D9D8EA}} \color[HTML]{000000} 27.2 & {\cellcolor[HTML]{023858}} \color[HTML]{F1F1F1} 58.9 & {\cellcolor[HTML]{549CC7}} \color[HTML]{F1F1F1} 40.8 & {\cellcolor[HTML]{8EB3D5}} \color[HTML]{000000} 35.8 & {\cellcolor[HTML]{ACC0DD}} \color[HTML]{000000} 32.7 & {\cellcolor[HTML]{DAD9EA}} \color[HTML]{000000} 26.8 \\
Betrayal & {\cellcolor[HTML]{F7F0F7}} \color[HTML]{000000} 21.0 & {\cellcolor[HTML]{FEF6FB}} \color[HTML]{000000} 19.2 & {\cellcolor[HTML]{D6D6E9}} \color[HTML]{000000} 27.7 & {\cellcolor[HTML]{E3E0EE}} \color[HTML]{000000} 25.4 & {\cellcolor[HTML]{FFF7FB}} \color[HTML]{000000} 15.5 & {\cellcolor[HTML]{91B5D6}} \color[HTML]{000000} 35.5 & {\cellcolor[HTML]{549CC7}} \color[HTML]{F1F1F1} 40.8 & {\cellcolor[HTML]{023858}} \color[HTML]{F1F1F1} 58.3 & {\cellcolor[HTML]{04649E}} \color[HTML]{F1F1F1} 50.6 & {\cellcolor[HTML]{8FB4D6}} \color[HTML]{000000} 35.7 & {\cellcolor[HTML]{AFC1DD}} \color[HTML]{000000} 32.5 \\
Subversion & {\cellcolor[HTML]{FFF7FB}} \color[HTML]{000000} 15.2 & {\cellcolor[HTML]{F1EBF5}} \color[HTML]{000000} 22.4 & {\cellcolor[HTML]{FEF6FB}} \color[HTML]{000000} 19.1 & {\cellcolor[HTML]{7EADD1}} \color[HTML]{F1F1F1} 37.4 & {\cellcolor[HTML]{FFF7FB}} \color[HTML]{000000} 17.5 & {\cellcolor[HTML]{C2CBE2}} \color[HTML]{000000} 30.2 & {\cellcolor[HTML]{8EB3D5}} \color[HTML]{000000} 35.8 & {\cellcolor[HTML]{04649E}} \color[HTML]{F1F1F1} 50.6 & {\cellcolor[HTML]{023858}} \color[HTML]{F1F1F1} 57.9 & {\cellcolor[HTML]{89B1D4}} \color[HTML]{000000} 36.2 & {\cellcolor[HTML]{BFC9E1}} \color[HTML]{000000} 30.7 \\
Degradation & {\cellcolor[HTML]{FFF7FB}} \color[HTML]{000000} 11.6 & {\cellcolor[HTML]{FFF7FB}} \color[HTML]{000000} 10.9 & {\cellcolor[HTML]{FFF7FB}} \color[HTML]{000000} 8.7 & {\cellcolor[HTML]{FFF7FB}} \color[HTML]{000000} 14.1 & {\cellcolor[HTML]{F1EBF5}} \color[HTML]{000000} 22.4 & {\cellcolor[HTML]{B5C4DF}} \color[HTML]{000000} 31.7 & {\cellcolor[HTML]{ACC0DD}} \color[HTML]{000000} 32.7 & {\cellcolor[HTML]{8FB4D6}} \color[HTML]{000000} 35.7 & {\cellcolor[HTML]{89B1D4}} \color[HTML]{000000} 36.2 & {\cellcolor[HTML]{157AB5}} \color[HTML]{F1F1F1} 46.5 & {\cellcolor[HTML]{D2D2E7}} \color[HTML]{000000} 28.5 \\
\hhline{~-----------}
Non-moral & {\cellcolor[HTML]{CED0E6}} \color[HTML]{000000} 28.8 & {\cellcolor[HTML]{DDDBEC}} \color[HTML]{000000} 26.4 & {\cellcolor[HTML]{D9D8EA}} \color[HTML]{000000} 27.0 & {\cellcolor[HTML]{D8D7E9}} \color[HTML]{000000} 27.3 & {\cellcolor[HTML]{D9D8EA}} \color[HTML]{000000} 27.2 & {\cellcolor[HTML]{C4CBE3}} \color[HTML]{000000} 30.0 & {\cellcolor[HTML]{DAD9EA}} \color[HTML]{000000} 26.9 & {\cellcolor[HTML]{AFC1DD}} \color[HTML]{000000} 32.5 & {\cellcolor[HTML]{BFC9E1}} \color[HTML]{000000} 30.7 & {\cellcolor[HTML]{D2D2E7}} \color[HTML]{000000} 28.5 & {\cellcolor[HTML]{BDC8E1}} \color[HTML]{000000} 30.8 \\
\end{tabular}}
\caption{Moral similarity for the MFTC training set with the supervised SimCSE approach. A darker color indicates higher similarity.}
\label{tab:moral-sim-MFTC}
\end{table}

First, we notice a high similarity along the diagonal, indicating that the moral embedding space consistently clusters data points annotated with the same label.
Further, the overall similarity between virtues and vices values (top-right and bottom-left quadrants) is visibly lower than the similarity between virtue-virtue \rebuttal{(top-left quadrant)} and vice-vice values \rebuttal{(bottom-right quadrant)}, which indicates that the model can clearly separate virtues and vices found in tweets.
Moreover, a significant similarity between opposing virtues and vices (e.g., \textit{fairness} and \textit{cheating}) can be observed, showing that the embedding space has learned relationships among corresponding virtues and vices.
Finally, the similarity between non-moral and moral values is modest, confirming that tweets labeled as \textit{non-moral} are spread throughout the embedding space, without forming any significant cluster.

\begin{figure*}[!htb]
    \centering
    \includegraphics[width=\linewidth]{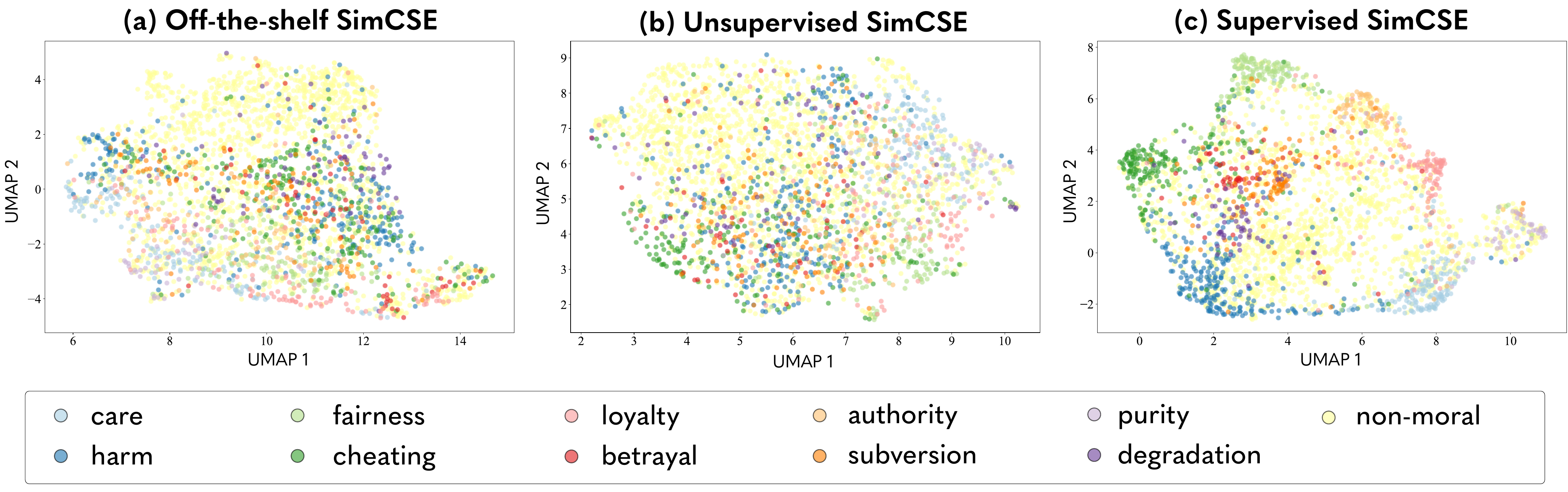}
    \caption{\rebuttal{UMAP plot of the \textbf{MFTC test set} data with off-the-shelf pre-trained SimCSE model (a, left), unsupervised SimCSE approach (b, middle), and supervised SimCSE approach (c, right).}}
    \label{fig:umap_moralD_after}
\end{figure*}

The results described above show the effectiveness of the training strategy described in Section~\ref{sec:method}. However, additional emergent results can be observed in Table~\ref{tab:moral-sim-MFTC}. For instance, on the diagonal, virtue values (top-left quadrant) have a higher similarity than vice values (bottom-right quadrant), showing that tweets labeled with virtue values are more consistently clustered.
Moreover, we observe that some elements have a high similarity despite not having been explicitly addressed by the training strategy, e.g., \textit{care-purity} and \textit{subversion-betrayal}. 

To further investigate these similarities, we tokenize and lemmatize the tweets labeled with these elements and inspect whether they share commonly used lemmas. We provide some insightful examples to better understand such similarities.
The word `god' appears consistently in tweets labeled with \textit{care} and \textit{purity}, hinting that the correlation is driven by common concerns of religion and care, especially in the context of the Sandy hurricane relief tweets.
The words `Obama' and `protest' are common for both \textit{betrayal} and \textit{subversion} tweets, showing how the correlation was driven by the political background behind tweets collected with the All Lives Matter and Black Lives Matter hashtags.

\rebuttal{Lastly, similar to Figure~\ref{fig:umap}, the moral similarity tables obtained with the off-the-shelf model and with the unsupervised SimCSE approach fail to produce meaningful similarities (see Appendix~\ref{app:intrinsic-moral-sim}).}

\subsection{Extrinsic Evaluation}
We present the results of \rebuttal{generalizability on the test set} and comparison to MFD2.0 dictionary.

\subsubsection{Generalizability on Test Set}

\rebuttal{Figure~\ref{fig:umap_moralD_after} shows the UMAP plot of the MFTC test set data mapped on the embedding spaces obtained with the three compared approaches.}
\rebuttal{First, we remark that the lower density of the plotted data with respect to Figure~\ref{fig:umap} is due to the smaller size of the test set compared to the training set.
Further, with the supervised SimCSE approach, we observe clear clusters corresponding to the MFT elements (similar to Figure~\ref{fig:umap}c). Instead, the UMAP plots resulting from the off-the-shelf model and the unsupervised approach show no distinguishable clusters.}

\begin{table}[!htb]
\centering 
\hspace{-0.68cm}
\resizebox{0.98\linewidth}{!}{
\scriptsize
\renewcommand{\tabcolsep}{1.3pt}
\renewcommand{\arraystretch}{1.4} 
\begin{tabular}{rccccc|ccccc|c}
 & \rot{Care} & \rot{Fairness} & \rot{Loyalty} & \rot{Authority} & \rot{Purity} & \rot{Harm} & \rot{Cheating} & \rot{Betrayal} & \rot{Subversion} & \rot{Degradation} &
 \rot{Non-moral}\\
Care & {\cellcolor[HTML]{023858}} \color[HTML]{F1F1F1} 75.2 & {\cellcolor[HTML]{DBDAEB}} \color[HTML]{000000} 26.7 & {\cellcolor[HTML]{4A98C5}} \color[HTML]{F1F1F1} 41.6 & {\cellcolor[HTML]{81AED2}} \color[HTML]{F1F1F1} 37.0 & {\cellcolor[HTML]{0568A3}} \color[HTML]{F1F1F1} 49.8 & {\cellcolor[HTML]{D2D2E7}} \color[HTML]{000000} 28.4 & {\cellcolor[HTML]{FFF7FB}} \color[HTML]{000000} 7.7 & {\cellcolor[HTML]{FBF3F9}} \color[HTML]{000000} 20.0 & {\cellcolor[HTML]{FFF7FB}} \color[HTML]{000000} 17.1 & {\cellcolor[HTML]{FFF7FB}} \color[HTML]{000000} 12.6 & {\cellcolor[HTML]{C8CDE4}} \color[HTML]{000000} 29.5 \\
Fairness & {\cellcolor[HTML]{DBDAEB}} \color[HTML]{000000} 26.7 & {\cellcolor[HTML]{023858}} \color[HTML]{F1F1F1} 72.0 & {\cellcolor[HTML]{D3D4E7}} \color[HTML]{000000} 28.1 & {\cellcolor[HTML]{4E9AC6}} \color[HTML]{F1F1F1} 41.3 & {\cellcolor[HTML]{BDC8E1}} \color[HTML]{000000} 30.8 & {\cellcolor[HTML]{FFF7FB}} \color[HTML]{000000} 15.6 & {\cellcolor[HTML]{94B6D7}} \color[HTML]{000000} 35.1 & {\cellcolor[HTML]{F2ECF5}} \color[HTML]{000000} 22.1 & {\cellcolor[HTML]{EAE6F1}} \color[HTML]{000000} 24.1 & {\cellcolor[HTML]{FFF7FB}} \color[HTML]{000000} 15.2 & {\cellcolor[HTML]{DCDAEB}} \color[HTML]{000000} 26.5 \\
Loyalty & {\cellcolor[HTML]{4A98C5}} \color[HTML]{F1F1F1} 41.6 & {\cellcolor[HTML]{D3D4E7}} \color[HTML]{000000} 28.1 & {\cellcolor[HTML]{023858}} \color[HTML]{F1F1F1} 60.8 & {\cellcolor[HTML]{79ABD0}} \color[HTML]{F1F1F1} 37.8 & {\cellcolor[HTML]{81AED2}} \color[HTML]{F1F1F1} 37.0 & {\cellcolor[HTML]{FFF7FB}} \color[HTML]{000000} 12.6 & {\cellcolor[HTML]{FFF7FB}} \color[HTML]{000000} 10.3 & {\cellcolor[HTML]{D9D8EA}} \color[HTML]{000000} 26.9 & {\cellcolor[HTML]{FBF4F9}} \color[HTML]{000000} 19.9 & {\cellcolor[HTML]{FFF7FB}} \color[HTML]{000000} 11.6 & {\cellcolor[HTML]{D6D6E9}} \color[HTML]{000000} 27.6 \\
Authority & {\cellcolor[HTML]{81AED2}} \color[HTML]{F1F1F1} 37.0 & {\cellcolor[HTML]{4E9AC6}} \color[HTML]{F1F1F1} 41.3 & {\cellcolor[HTML]{79ABD0}} \color[HTML]{F1F1F1} 37.8 & {\cellcolor[HTML]{023858}} \color[HTML]{F1F1F1} 62.9 & {\cellcolor[HTML]{4094C3}} \color[HTML]{F1F1F1} 42.4 & {\cellcolor[HTML]{FFF7FB}} \color[HTML]{000000} 14.7 & {\cellcolor[HTML]{FFF7FB}} \color[HTML]{000000} 16.2 & {\cellcolor[HTML]{ECE7F2}} \color[HTML]{000000} 23.9 & {\cellcolor[HTML]{9EBAD9}} \color[HTML]{000000} 34.3 & {\cellcolor[HTML]{FEF6FB}} \color[HTML]{000000} 19.1 & {\cellcolor[HTML]{D5D5E8}} \color[HTML]{000000} 27.7 \\
Purity & {\cellcolor[HTML]{0568A3}} \color[HTML]{F1F1F1} 49.8 & {\cellcolor[HTML]{BDC8E1}} \color[HTML]{000000} 30.8 & {\cellcolor[HTML]{81AED2}} \color[HTML]{F1F1F1} 37.0 & {\cellcolor[HTML]{4094C3}} \color[HTML]{F1F1F1} 42.4 & {\cellcolor[HTML]{023858}} \color[HTML]{F1F1F1} 75.5 & {\cellcolor[HTML]{FFF7FB}} \color[HTML]{000000} 15.1 & {\cellcolor[HTML]{FFF7FB}} \color[HTML]{000000} 6.3 & {\cellcolor[HTML]{FFF7FB}} \color[HTML]{000000} 13.9 & {\cellcolor[HTML]{FFF7FB}} \color[HTML]{000000} 17.6 & {\cellcolor[HTML]{FFF7FB}} \color[HTML]{000000} 18.7 & {\cellcolor[HTML]{D6D6E9}} \color[HTML]{000000} 27.6 \\
\hhline{~-----------}
Harm & {\cellcolor[HTML]{D2D2E7}} \color[HTML]{000000} 28.4 & {\cellcolor[HTML]{FFF7FB}} \color[HTML]{000000} 15.6 & {\cellcolor[HTML]{FFF7FB}} \color[HTML]{000000} 12.6 & {\cellcolor[HTML]{FFF7FB}} \color[HTML]{000000} 14.7 & {\cellcolor[HTML]{FFF7FB}} \color[HTML]{000000} 15.1 & {\cellcolor[HTML]{045E93}} \color[HTML]{F1F1F1} 52.1 & {\cellcolor[HTML]{DDDBEC}} \color[HTML]{000000} 26.4 & {\cellcolor[HTML]{96B6D7}} \color[HTML]{000000} 35.0 & {\cellcolor[HTML]{B1C2DE}} \color[HTML]{000000} 32.2 & {\cellcolor[HTML]{AFC1DD}} \color[HTML]{000000} 32.5 & {\cellcolor[HTML]{C2CBE2}} \color[HTML]{000000} 30.2 \\
Cheating & {\cellcolor[HTML]{FFF7FB}} \color[HTML]{000000} 7.7 & {\cellcolor[HTML]{94B6D7}} \color[HTML]{000000} 35.1 & {\cellcolor[HTML]{FFF7FB}} \color[HTML]{000000} 10.3 & {\cellcolor[HTML]{FFF7FB}} \color[HTML]{000000} 16.2 & {\cellcolor[HTML]{FFF7FB}} \color[HTML]{000000} 6.3 & {\cellcolor[HTML]{DDDBEC}} \color[HTML]{000000} 26.4 & {\cellcolor[HTML]{034267}} \color[HTML]{F1F1F1} 56.4 & {\cellcolor[HTML]{4C99C5}} \color[HTML]{F1F1F1} 41.5 & {\cellcolor[HTML]{9CB9D9}} \color[HTML]{000000} 34.5 & {\cellcolor[HTML]{A5BDDB}} \color[HTML]{000000} 33.5 & {\cellcolor[HTML]{DEDCEC}} \color[HTML]{000000} 26.2 \\
Betrayal & {\cellcolor[HTML]{FBF3F9}} \color[HTML]{000000} 20.0 & {\cellcolor[HTML]{F2ECF5}} \color[HTML]{000000} 22.1 & {\cellcolor[HTML]{D9D8EA}} \color[HTML]{000000} 26.9 & {\cellcolor[HTML]{ECE7F2}} \color[HTML]{000000} 23.9 & {\cellcolor[HTML]{FFF7FB}} \color[HTML]{000000} 13.9 & {\cellcolor[HTML]{96B6D7}} \color[HTML]{000000} 35.0 & {\cellcolor[HTML]{4C99C5}} \color[HTML]{F1F1F1} 41.5 & {\cellcolor[HTML]{023E62}} \color[HTML]{F1F1F1} 56.8 & {\cellcolor[HTML]{1077B4}} \color[HTML]{F1F1F1} 46.9 & {\cellcolor[HTML]{67A4CC}} \color[HTML]{F1F1F1} 39.3 & {\cellcolor[HTML]{B4C4DF}} \color[HTML]{000000} 31.8 \\
Subversion & {\cellcolor[HTML]{FFF7FB}} \color[HTML]{000000} 17.1 & {\cellcolor[HTML]{EAE6F1}} \color[HTML]{000000} 24.1 & {\cellcolor[HTML]{FBF4F9}} \color[HTML]{000000} 19.9 & {\cellcolor[HTML]{9EBAD9}} \color[HTML]{000000} 34.3 & {\cellcolor[HTML]{FFF7FB}} \color[HTML]{000000} 17.6 & {\cellcolor[HTML]{B1C2DE}} \color[HTML]{000000} 32.2 & {\cellcolor[HTML]{9CB9D9}} \color[HTML]{000000} 34.5 & {\cellcolor[HTML]{1077B4}} \color[HTML]{F1F1F1} 46.9 & {\cellcolor[HTML]{045F95}} \color[HTML]{F1F1F1} 51.8 & {\cellcolor[HTML]{5A9EC9}} \color[HTML]{F1F1F1} 40.4 & {\cellcolor[HTML]{C1CAE2}} \color[HTML]{000000} 30.4 \\
Degradation & {\cellcolor[HTML]{FFF7FB}} \color[HTML]{000000} 12.6 & {\cellcolor[HTML]{FFF7FB}} \color[HTML]{000000} 15.2 & {\cellcolor[HTML]{FFF7FB}} \color[HTML]{000000} 11.6 & {\cellcolor[HTML]{FEF6FB}} \color[HTML]{000000} 19.1 & {\cellcolor[HTML]{FFF7FB}} \color[HTML]{000000} 18.7 & {\cellcolor[HTML]{AFC1DD}} \color[HTML]{000000} 32.5 & {\cellcolor[HTML]{A5BDDB}} \color[HTML]{000000} 33.5 & {\cellcolor[HTML]{67A4CC}} \color[HTML]{F1F1F1} 39.3 & {\cellcolor[HTML]{5A9EC9}} \color[HTML]{F1F1F1} 40.4 & {\cellcolor[HTML]{157AB5}} \color[HTML]{F1F1F1} 46.5 & {\cellcolor[HTML]{C6CCE3}} \color[HTML]{000000} 29.7 \\
\hhline{~-----------}
Non-Moral & {\cellcolor[HTML]{C8CDE4}} \color[HTML]{000000} 29.5 & {\cellcolor[HTML]{DCDAEB}} \color[HTML]{000000} 26.5 & {\cellcolor[HTML]{D6D6E9}} \color[HTML]{000000} 27.6 & {\cellcolor[HTML]{D5D5E8}} \color[HTML]{000000} 27.7 & {\cellcolor[HTML]{D6D6E9}} \color[HTML]{000000} 27.6 & {\cellcolor[HTML]{C2CBE2}} \color[HTML]{000000} 30.2 & {\cellcolor[HTML]{DEDCEC}} \color[HTML]{000000} 26.2 & {\cellcolor[HTML]{B4C4DF}} \color[HTML]{000000} 31.8 & {\cellcolor[HTML]{C1CAE2}} \color[HTML]{000000} 30.4 & {\cellcolor[HTML]{C6CCE3}} \color[HTML]{000000} 29.7 & {\cellcolor[HTML]{BCC7E1}} \color[HTML]{000000} 30.9 \\
\end{tabular}}
\caption{Moral similarity for MFTC test set with supervised SimCSE. Darker the cell higher the similarity.}
\label{tab:moral-sim-MFTC-test}
\end{table}

\rebuttal{To quantitatively investigate the relationships among the MFT elements, we show in Table~\ref{tab:moral-sim-MFTC-test} the moral similarity for the MFTC test set with the supervised SimCSE approach.
These results are in line with Table~\ref{tab:moral-sim-MFTC}, and show that the distribution of the MFT elements learned in the training set is consistent with the data in the test set.}

\subsubsection{Comparison to MFD2.0}

We present the results of the clustering of the MFD2.0 words based on the \rebuttal{three compared approaches (as described in Section~\ref{sec:experimental-setup-extrinsic})}. We further inspect the best-performing approach through the moral similarity evaluation of the MFD2.0 words.

\paragraph{Clustering}
\label{sec:result-clustering}

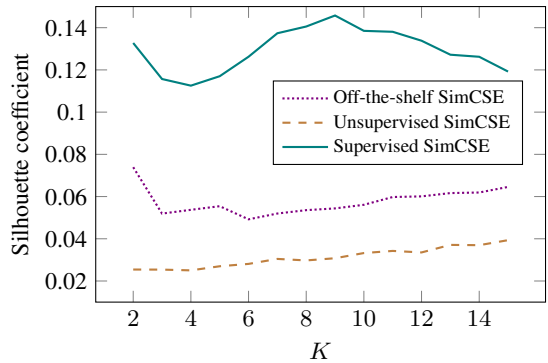
\begin{figure}[!htb]
\small
\begin{tikzpicture}
    \begin{axis}[
        width=7.5cm,
        height=5.5cm,
        xlabel = {$K$},
        ylabel = {Silhouette coefficient},
        ymin = 0.01, ymax = 0.15,
        xtick={2,4,6,8,10,12,14},
        ytick={0.02,0.04,0.06,0.08,0.1,0.12,0.14},
        yticklabels={0.02,0.04,0.06,0.08,0.1,0.12,0.14},
        legend style={nodes={scale=0.8},at={(0.95,0.75)}},
        legend cell align={left}
    ]
    \addplot [violet, thick, densely dotted] table {tikz/silhouette/sup_base.dat};
    \addlegendentry{\rebuttal{Off-the-shelf SimCSE}};
    \addplot [brown, thick, dashed] table {tikz/silhouette/unsup.dat};
    \addlegendentry{Unsupervised SimCSE};
    \addplot [teal, thick] table {tikz/silhouette/sup.dat};
    \addlegendentry{Supervised SimCSE};
    \end{axis}
\end{tikzpicture}
\caption{Silhouette coefficients for $K$ ranging from 2 to 15 for the \rebuttal{three compared approaches}.}
\label{fig:clustering-k}
\end{figure}

Figure~\ref{fig:clustering-k} shows the silhouette coefficient for K-means clustering with $K$ ranging from 2 to 15 for the three compared approaches. 

We observe that the supervised SimCSE approach performs best, with a silhouette coefficient that peaks at $K=9$, close to the total number of MFT elements (10). Instead, the \rebuttal{off-the-shelf model} peaks at $K=2$, aligning with previous research results that show that the pre-trained embedding spaces contain an intuitive distinction between do's and don'ts \cite{nmi}.
Further, we observe low silhouette coefficients due to the high dimensionality of the embedding space.

Table~\ref{tab:clustering-purity} shows purity and silhouette coefficients for $K=\hat{K}$ (the $K$ that leads to the highest silhouette coefficient) and $K=10$.
The supervised SimCSE approach achieves the highest purity score for both $K=\hat{K}$ and $K=10$, resulting in a purity of 0.71 in both cases. This result shows that the resulting embedding space allows for a coherent clustering of the MFD2.0 words, proving consistent with an independently generated MFT dictionary.

\begin{table}[!htb]
\centering
\small
\begin{tabular}{@{}c|lccc@{}}
\toprule
& \textbf{Approach} & \textbf{$K$} & \textbf{Purity} & \textbf{Silhouette} \\ \midrule
& \rebuttal{Off-the-shelf SimCSE}   & 2  & 0.30  & 0.07  \\
& Unsupervised SimCSE & 15 & 0.51  & 0.04      \\
\multirow{-3}{*}{\begin{sideways}$K=\hat{K}$\end{sideways}} & Supervised SimCSE   & 9  & \textbf{0.71}  & \textbf{0.15}      \\
\midrule
& \rebuttal{Off-the-shelf SimCSE}   & 10 & 0.56  & 0.06  \\
& Unsupervised SimCSE  & 10 & 0.45  & 0.03      \\
\multirow{-3}{*}{\begin{sideways}$K=10$\end{sideways}} & Supervised SimCSE   & 10 & \textbf{0.71}  & \textbf{0.14}      \\ 
\bottomrule
\end{tabular}
\caption{Purity and Silhouette coefficients for $K=\hat{K}$ and $K=10$. The best scores are highlighted in bold.}
\label{tab:clustering-purity}
\end{table}

\paragraph{Moral Similarity \rebuttal{(MFD2.0)}}
\label{sec:result-moral-sim}
We further investigate the consistency between the supervised SimCSE embedding space approach and MFD2.0.
Table~\ref{tab:moral-sim-mfd2} shows the moral similarity between the MFT elements, calculated with the supervised SimCSE embedding space representation of MFD2.0 words.

\begin{table}[!htb]
\centering
\scriptsize
\hspace{-0.7cm}
\renewcommand{\tabcolsep}{1.5pt}
\renewcommand{\arraystretch}{1.4} 
\begin{tabular}{rccccc|ccccc}
 & \rot{Care} & \rot{Fairness} & \rot{Loyalty} & \rot{Authority} & \rot{Purity} & \rot{Harm} & \rot{Cheating} & \rot{Betrayal} & \rot{Subversion} & \rot{Degradation} \\
Care & {\cellcolor[HTML]{023858}} \color[HTML]{F1F1F1} 57.8 & {\cellcolor[HTML]{BFC9E1}} \color[HTML]{000000} 30.7 & {\cellcolor[HTML]{84B0D3}} \color[HTML]{F1F1F1} 36.7 & {\cellcolor[HTML]{B0C2DE}} \color[HTML]{000000} 32.4 & {\cellcolor[HTML]{67A4CC}} \color[HTML]{F1F1F1} 39.4 & {\cellcolor[HTML]{C4CBE3}} \color[HTML]{000000} 30.1 & {\cellcolor[HTML]{FFF7FB}} \color[HTML]{000000} 18.9 & {\cellcolor[HTML]{EEE9F3}} \color[HTML]{000000} 23.2 & {\cellcolor[HTML]{FDF5FA}} \color[HTML]{000000} 19.4 & {\cellcolor[HTML]{F1EBF5}} \color[HTML]{000000} 22.4 \\
Fairness & {\cellcolor[HTML]{BFC9E1}} \color[HTML]{000000} 30.7 & {\cellcolor[HTML]{056FAE}} \color[HTML]{F1F1F1} 48.3 & {\cellcolor[HTML]{ABBFDC}} \color[HTML]{000000} 33.0 & {\cellcolor[HTML]{7DACD1}} \color[HTML]{F1F1F1} 37.5 & {\cellcolor[HTML]{AFC1DD}} \color[HTML]{000000} 32.5 & {\cellcolor[HTML]{E5E1EF}} \color[HTML]{000000} 25.1 & {\cellcolor[HTML]{C2CBE2}} \color[HTML]{000000} 30.3 & {\cellcolor[HTML]{D5D5E8}} \color[HTML]{000000} 27.8 & {\cellcolor[HTML]{D7D6E9}} \color[HTML]{000000} 27.5 & {\cellcolor[HTML]{F2ECF5}} \color[HTML]{000000} 22.2 \\
Loyalty & {\cellcolor[HTML]{84B0D3}} \color[HTML]{F1F1F1} 36.7 & {\cellcolor[HTML]{ABBFDC}} \color[HTML]{000000} 33.0 & {\cellcolor[HTML]{04639B}} \color[HTML]{F1F1F1} 50.9 & {\cellcolor[HTML]{8EB3D5}} \color[HTML]{000000} 35.8 & {\cellcolor[HTML]{75A9CF}} \color[HTML]{F1F1F1} 38.3 & {\cellcolor[HTML]{DCDAEB}} \color[HTML]{000000} 26.5 & {\cellcolor[HTML]{E7E3F0}} \color[HTML]{000000} 24.6 & {\cellcolor[HTML]{A7BDDB}} \color[HTML]{000000} 33.4 & {\cellcolor[HTML]{B4C4DF}} \color[HTML]{000000} 31.9 & {\cellcolor[HTML]{D7D6E9}} \color[HTML]{000000} 27.4 \\
Authority & {\cellcolor[HTML]{B0C2DE}} \color[HTML]{000000} 32.4 & {\cellcolor[HTML]{7DACD1}} \color[HTML]{F1F1F1} 37.5 & {\cellcolor[HTML]{8EB3D5}} \color[HTML]{000000} 35.8 & {\cellcolor[HTML]{056FAE}} \color[HTML]{F1F1F1} 48.2 & {\cellcolor[HTML]{60A1CA}} \color[HTML]{F1F1F1} 40.0 & {\cellcolor[HTML]{DFDDEC}} \color[HTML]{000000} 26.1 & {\cellcolor[HTML]{E2DFEE}} \color[HTML]{000000} 25.5 & {\cellcolor[HTML]{B9C6E0}} \color[HTML]{000000} 31.3 & {\cellcolor[HTML]{88B1D4}} \color[HTML]{000000} 36.4 & {\cellcolor[HTML]{D7D6E9}} \color[HTML]{000000} 27.4 \\
Purity & {\cellcolor[HTML]{67A4CC}} \color[HTML]{F1F1F1} 39.4 & {\cellcolor[HTML]{AFC1DD}} \color[HTML]{000000} 32.5 & {\cellcolor[HTML]{75A9CF}} \color[HTML]{F1F1F1} 38.3 & {\cellcolor[HTML]{60A1CA}} \color[HTML]{F1F1F1} 40.0 & {\cellcolor[HTML]{023B5D}} \color[HTML]{F1F1F1} 57.2 & {\cellcolor[HTML]{D9D8EA}} \color[HTML]{000000} 27.0 & {\cellcolor[HTML]{F6EFF7}} \color[HTML]{000000} 21.2 & {\cellcolor[HTML]{D7D6E9}} \color[HTML]{000000} 27.4 & {\cellcolor[HTML]{BFC9E1}} \color[HTML]{000000} 30.7 & {\cellcolor[HTML]{97B7D7}} \color[HTML]{000000} 35.0 \\
\hhline{~----------}
Harm & {\cellcolor[HTML]{C4CBE3}} \color[HTML]{000000} 30.1 & {\cellcolor[HTML]{E5E1EF}} \color[HTML]{000000} 25.1 & {\cellcolor[HTML]{DCDAEB}} \color[HTML]{000000} 26.5 & {\cellcolor[HTML]{DFDDEC}} \color[HTML]{000000} 26.1 & {\cellcolor[HTML]{D9D8EA}} \color[HTML]{000000} 27.0 & {\cellcolor[HTML]{034267}} \color[HTML]{F1F1F1} 56.4 & {\cellcolor[HTML]{8CB3D5}} \color[HTML]{000000} 35.9 & {\cellcolor[HTML]{91B5D6}} \color[HTML]{000000} 35.6 & {\cellcolor[HTML]{A7BDDB}} \color[HTML]{000000} 33.5 & {\cellcolor[HTML]{4697C4}} \color[HTML]{F1F1F1} 41.8 \\
Cheating & {\cellcolor[HTML]{FFF7FB}} \color[HTML]{000000} 18.9 & {\cellcolor[HTML]{C2CBE2}} \color[HTML]{000000} 30.3 & {\cellcolor[HTML]{E7E3F0}} \color[HTML]{000000} 24.6 & {\cellcolor[HTML]{E2DFEE}} \color[HTML]{000000} 25.5 & {\cellcolor[HTML]{F6EFF7}} \color[HTML]{000000} 21.2 & {\cellcolor[HTML]{8CB3D5}} \color[HTML]{000000} 35.9 & {\cellcolor[HTML]{045C90}} \color[HTML]{F1F1F1} 52.4 & {\cellcolor[HTML]{1B7EB7}} \color[HTML]{F1F1F1} 45.9 & {\cellcolor[HTML]{549CC7}} \color[HTML]{F1F1F1} 40.9 & {\cellcolor[HTML]{67A4CC}} \color[HTML]{F1F1F1} 39.3 \\
Betrayal & {\cellcolor[HTML]{EEE9F3}} \color[HTML]{000000} 23.2 & {\cellcolor[HTML]{D5D5E8}} \color[HTML]{000000} 27.8 & {\cellcolor[HTML]{A7BDDB}} \color[HTML]{000000} 33.4 & {\cellcolor[HTML]{B9C6E0}} \color[HTML]{000000} 31.3 & {\cellcolor[HTML]{D7D6E9}} \color[HTML]{000000} 27.4 & {\cellcolor[HTML]{91B5D6}} \color[HTML]{000000} 35.6 & {\cellcolor[HTML]{1B7EB7}} \color[HTML]{F1F1F1} 45.9 & {\cellcolor[HTML]{034C78}} \color[HTML]{F1F1F1} 54.9 & {\cellcolor[HTML]{04629A}} \color[HTML]{F1F1F1} 51.0 & {\cellcolor[HTML]{67A4CC}} \color[HTML]{F1F1F1} 39.3 \\
Subversion & {\cellcolor[HTML]{FDF5FA}} \color[HTML]{000000} 19.4 & {\cellcolor[HTML]{D7D6E9}} \color[HTML]{000000} 27.5 & {\cellcolor[HTML]{B4C4DF}} \color[HTML]{000000} 31.9 & {\cellcolor[HTML]{88B1D4}} \color[HTML]{000000} 36.4 & {\cellcolor[HTML]{BFC9E1}} \color[HTML]{000000} 30.7 & {\cellcolor[HTML]{A7BDDB}} \color[HTML]{000000} 33.5 & {\cellcolor[HTML]{549CC7}} \color[HTML]{F1F1F1} 40.9 & {\cellcolor[HTML]{04629A}} \color[HTML]{F1F1F1} 51.0 & {\cellcolor[HTML]{034165}} \color[HTML]{F1F1F1} 56.5 & {\cellcolor[HTML]{509AC6}} \color[HTML]{F1F1F1} 41.1 \\
Degradation & {\cellcolor[HTML]{F1EBF5}} \color[HTML]{000000} 22.4 & {\cellcolor[HTML]{F2ECF5}} \color[HTML]{000000} 22.2 & {\cellcolor[HTML]{D7D6E9}} \color[HTML]{000000} 27.4 & {\cellcolor[HTML]{D7D6E9}} \color[HTML]{000000} 27.4 & {\cellcolor[HTML]{97B7D7}} \color[HTML]{000000} 35.0 & {\cellcolor[HTML]{4697C4}} \color[HTML]{F1F1F1} 41.8 & {\cellcolor[HTML]{67A4CC}} \color[HTML]{F1F1F1} 39.3 & {\cellcolor[HTML]{67A4CC}} \color[HTML]{F1F1F1} 39.3 & {\cellcolor[HTML]{509AC6}} \color[HTML]{F1F1F1} 41.1 & {\cellcolor[HTML]{045382}} \color[HTML]{F1F1F1} 53.9 \\
\end{tabular}
\caption{Moral similarity for MFD2.0 with supervised SimCSE. Darker the cell higher the similarity.}
\label{tab:moral-sim-mfd2}
\end{table}

The high similarity along the diagonal indicates that MFD2.0 words that represent the same moral value are closer in embedding space with respect to words that represent different moral values.
Further, we notice parallels with Table~\ref{tab:moral-sim-MFTC}. That is, 
\begin{enumerate*}[label=(\arabic*)]
    \item the similarity between virtues and virtues (top-left quadrant) and vices and vices (bottom-right quadrant) is greater than the similarity between virtues and vices (top-right and bottom-left quadrants), and
    \item there is a noticeable similarity between corresponding virtues and vices (e.g., \textit{authority} and \textit{subversion}).
\end{enumerate*}
These results confirm that the supervised SimCSE approach generates moral embeddings that align with an independently generated MFT dictionary, \rebuttal{whereas the off-the-shelf and unsupervised approaches fail to do so (Appendix~\ref{appendix:MFD2.0}).}

\section{Related Works}
We review previous research on methods for detecting moral values and existing moral datasets.

\subsection{Detecting Moral Values in Text}

Traditionally, value lexicons---sets of words descriptive of each moral element---have been used to detect morality through text similarity \cite{Bahngat2020,Pavan2020MoralityText}. \citet{Graham2009} developed the Moral Foundations Dictionary (MFD), which has been extended manually \cite{Frimer_2019} and via semi-automated methods \cite{rezapour-etal-2019-enhancing, Araque_2020, kobbe-etal-2020-exploring, Hopp2020TheEM}.
However, word-level lexicons are limited by the ambiguity of natural language and the restricted range of lemmas, which can be solved by projecting the MFD lexicon on knowledge graphs that link moral entities and concepts \cite{hulpus-etal-2020-knowledge,asprino-etal-2022-uncovering}.
Other methods instead use the supervised classification paradigm \cite{Lin2017AcquiringBK,johnson-goldwasser-2018-classification,Hoover2020}, exploiting an annotated dataset to train a classifier. In particular, BERT-based models have been successfully used on datasets annotated with the MFT taxonomy \cite{kobbe-etal-2020-exploring,alshomary-etal-2022-moral,liscio-etal-2022-cross,Huang2022LearningWeighting,Bulla2023}.

Similar to our work, \citet{moralEmbeddingFloyd} map text onto a 10-dimensional space (corresponding to the MFT elements) where the position of a word in each dimension is determined by the moral valence that FrameAxis (an MFT-based lexicon \cite{frameAxis}) attributes to the word for the corresponding MFT element. 
Our work differs in that we use state-of-the-art pre-trained 1024-dimensional sentence embeddings that have been shown to be more effective at capturing semantic similarity compared to lexicon-based approaches.

\subsection{Datasets with Moral Content}
Besides the MFTC, other datasets based on different moral value taxonomies have been collected for NLP applications. \camera{The Schwartz value theory \cite{Schwartz2012d} is another commonly used taxonomy, composed of 20 values that form a continuum of meaning in a circumplex.} \citet{kiesel-etal-2022-identifying} presented a dataset of 5,270 arguments labeled with the Schwartz values and extended it to over 9K arguments for the SemEval-2023 Task 4 \cite{kiesel:2023a}.
\citet{Qiu_Zhao_Li_Lu_Peng_Gao_Zhu_2022} collected a dataset of dialogues in different social scenarios, also annotated with the Schwartz values. 
\citet{NEURIPS2022_b654d615} proposed MoralExceptQA, the novel challenge and dataset on moral exception question answering.
Finally, \citet{hendrycksEthics} introduced a dataset with contextualized scenarios about commonsense moral intuitions. 
We opted for MFT and MFTC due to the strong psychological background and the availability of a large annotated dataset.

\section{Conclusions and Future Work}
AI agents ought to recognize the diversity and nuances of human moral perspectives. To this end, we propose a method to generate a pluralist moral sentence embedding space with a state-of-the-art contrastive learning approach and focus on its evaluation. 
First, we perform an intrinsic evaluation to \rebuttal{evaluate the significance of label information for distinguishing among the different elements of pluralist morality.}
Our results show that a pluralist approach to morality cannot be simply learned through self-supervised learning, but human labels are essential.
Then, we demonstrate that the embedding space trained through label supervision is aligned with \rebuttal{externally sourced data such as an} independently created lexicon of words that are descriptive of a pluralist approach to morality. 

Our investigation opens avenues for incorporating a pluralist approach to morality in language models, overcoming a simplistic, binary interpretation, i.e., simply judging a situation as morally right or wrong. 
\rebuttal{Pluralist moral embeddings can be used in a variety of applications,} e.g., recognizing moral rhetoric from diverse social issues such as abortion and terrorism \cite{10.1177/0894439313506837}, generating morally-aligned language \cite{ammanabrolu-etal-2022-aligning,lorandi-belz-2023-control}, \camera{measuring disagreement in online discussions \cite{shortall2022reason,van-der-meer-etal-2023-differences},} and investigating the context specificity of moral judgment \cite{liscio2022values,Liscio2023} or the cultural influences on moral norms \cite{ramezani-xu-2023-knowledge}.
\camera{Furthermore, the detection of pluralist morality could be extended with Hybrid Intelligence approaches \cite{akata2020research}, aiming at devising AI systems that combine human and artificial intelligence by design (e.g., \citet{van2022hyena,siebert2022estimating}).}

Our experiments are limited to one dataset and one approach to moral pluralism. \rebuttal{However, our experimental setup can be extended to other corpora to assess the generalizability to other approaches to pluralist morality.}
\camera{For instance, a comparative analysis would reveal differences between discrete and fuzzy approaches to moral pluralism, e.g., by comparing the MFT and the Schwartz value theory \cite{Schwartz2012d}.}
Similarly, we chose SimCSE due to its proven efficacy, but additional CL approaches could extend our work, e.g., by incorporating label embeddings in the training procedure \cite{zhang-etal-2022-label} or by exploiting adversarial examples to improve generalizability \cite{zhan-etal-2023-contrastive}. 
\camera{Finally}, the MFTC was annotated by multiple annotators and we used the majority agreement to train the moral embedding space. To better reflect the subjective nature of morality, an avenue for future work is to employ all annotations, incorporating annotators' (dis)agreement through a perspectivist approach \cite{10.1613/jair.1.12752,cabitza2023toward}.

\section{Ethical Considerations and Limitations}
\label{sec:ethical-considerations}
Morally-charged content poses a significant challenge for language models \cite{NEURIPS2022_b654d615}. 
This is particularly problematic when models trained to discern descriptive ethics (i.e., understand how humans reason about moral judgments) are used for normative ethics, (i.e., to make moral judgments such as religious prescriptions and medical advice) \cite{talat-etal-2022-machine}. For this reason, in this work, we limit ourselves to descriptive ethics. Further, the usage of our embedding space in highly sensitive domains, such as the legal field, requires additional cautious deliberation \cite{leins-etal-2020-give}.

An additional challenge is introduced by the \textit{dual-use} problem \cite{hovy-spruit-2016-social}, that is when a system developed for a certain purpose leads to unintended negative consequences in another application. For instance, since liberals and conservatives rely on different moral foundations \cite{Graham2009}, the moral embedding space can be misused to identify and discriminate against people with certain political standpoints. 

Next, we recognize the limitations regarding the dataset we use, the MFTC. First of all, the MFTC is composed of English tweets about US-centric topics, thus perpetuating Western biases \cite{Mehrabi2021}.
Post-hoc debiasing techniques \cite{liang-etal-2020-towards} can be applied to the current moral embedding space, preventing the need for re-training with large amounts of additional data.
However, our method and evaluation procedure can be applied to larger and culturally diverse datasets as well.
Then, the MFTC annotation procedure resulted in a low annotator agreement, which is to be expected in such a subjective annotation task \cite{Hoover2020}. Choosing the majority label as the true label reinforces the domination of the majority, suppressing the minority views. Employing a perspectivist approach, using all the annotations when training, can improve the representativity of the embedding space \cite{cabitza2023toward}. 

Finally, we recognize concerns on the evaluation procedure. 
First, the MFT dictionary (MFD2.0) is based on the WEIRD (Western, Educated, Industrialized, Rich, Democratic) sample. Dictionaries created from more diverse samples could reveal new strengths and weaknesses of the embedding space.
Second, we used UMAP to easily visualize the embedding space and the effect of the training. Additional investigation is required for a detailed geometric analysis of the embedding space.



\section*{Acknowledgements}

This research was partially funded by the Netherlands Organisation for Scientific Research (NWO) through the Hybrid Intelligence Centre via the Zwaartekracht grant (024.004.022).

\bibliographystyle{acl_natbib}

\clearpage
\appendix

\renewcommand{\thefigure}{A\arabic{figure}}
\setcounter{figure}{0}
\renewcommand{\thetable}{A\arabic{table}}
\setcounter{table}{0}

\section{Experimental Details}
\label{sec:appendix1}
For the sake of reproducibility, we share further details on our experimental procedure. \camera{The trained models are available online \cite{4TUmodels}}.

\subsection{SimCSE Contrastive Losses}
\label{app:cl-loss}

\rebuttal{
We present the SimCSE contrastive losses as introduced by \citet{gao-etal-2021-simcse}.
For unsupervised SimCSE, we take a collection of sentences $\{x_i\}^{m}_{i=1}$, and uses $x_i^+ = x_i$. It constructs a positive pair for each input $x_i$ by encoding the input twice using
different dropout masks, $z$ and $z^{\prime}$. We denote $\mathbf{h}_i^z = f_{\theta}(x_i, z)$, where $z$ is a random mask for dropout. Note that in the standard transformer models, there are dropout masks placed on fully-connected layers. The training objective for the unsupervised SimCSE approach is the following:
\begin{equation*}
\ell_i = -\log \frac{e^{\operatorname{sim}\left(\mathbf{h}_i^{z_i}, \mathbf{h}_i^{z_i^{\prime}}\right) / \tau}}{\sum_{j=1}^N e^{\operatorname{sim}\left(\mathbf{h}_i^{z_i}, \mathbf{h}_j^{z_j^{\prime}}\right) / \tau}},
\end{equation*}
}

\rebuttal{For supervised SimCSE, instead of using dropout, it takes predefined positive and negative instances, $x_i^+$ and $x_i^-$ respectively. The training objective for the supervised SimCSE approach is the following: 
\begin{equation*}
\ell_i = -\log \frac{e^{\operatorname{sim}\left(\mathbf{h}_i, \mathbf{h}_i^{+}\right) / \tau}}{\sum_{j=1}^N\left(e^{\operatorname{sim}\left(\mathbf{h}_i, \mathbf{h}_j^{+}\right) / \tau}+e^{\operatorname{sim}\left(\mathbf{h}_i, \mathbf{h}_j^{-}\right) / \tau}\right)}
\end{equation*}
}

\subsection{Data Processing}
\label{app:data-process}
We preprocess the tweets by removing URLs, emails, usernames, and mentions. Next, we employ the Ekphrasis package\footnote{\url{https://github.com/cbaziotis/ekphrasis}} to correct common spelling mistakes and unpack contractions. Finally, emojis are transformed into their respective words using the Python Emoji package\footnote{\url{https://pypi.org/project/emoji/}}. 
Moreover, there are some independent tweets with duplicated content, in some cases with different labels. We reduced repeated instances of distinct tweet annotations to one instance by applying a majority vote. The final unsupervised SimCSE training set consists of 29,147 triples (i.e., the size of the training set). The final supervised SimCSE training set consists of 5,304 triples, due to the large number of \textit{non-moral} labels (Table~\ref{tab:label_distr}) that did not appear in any triple.

\subsection{Hyperparameters}
\label{app:hyperparams}
To select the most optimal combination of hyperparameters for SimCSE, we perform a grid search based on the $F_1$-scores of the classification result, \rebuttal{which is further discussed in Appendix \ref{sec:classification}}. Table \ref{tab:hyperparams-supervised} and Table \ref{tab:hyperparams-unsupervised} show the hyperparameters that were compared, highlighting in bold the best-performing option. We used these hyperparameters for every experiment in this paper for consistency. If a parameter is not present in the table, the default value supplied by the framework\footnote{\url{https://github.com/princeton-nlp/SimCSE}} was used. 

\begin{table}[!htb]
\small
\centering
\begin{tabular}{@{}ll@{}}
\toprule
\textbf{Hyperparameters} & \textbf{Options}              \\ \midrule
Model name               & sup-simcse-bert-large-uncased \\
Max Sequence Length      & \textbf{64}, 128                   \\
Epochs                   & \textbf{2}, 3, 5                       \\
Batch Size               & 16, \textbf{32}                    \\
Learning Rate            & $5\times 10^{-5}$\\
Temperature              & 0.01, 0.05, \textbf{ 0.1} \\ 
Pooler              & \textbf{cls} \\ \bottomrule
\end{tabular}
\caption{Hyperparameters tested for training SimCSE with the supervised approach.}
\label{tab:hyperparams-supervised}
\end{table}

\begin{table}[!htb]
\small
\centering
\begin{tabular}{@{}ll@{}}
\toprule
\textbf{Hyperparameters} & \textbf{Options}                \\ \midrule
Model name               & unsup-simcse-bert-large-uncased \\
Max Sequence Length      & \textbf{64}, 128                     \\
Epochs                   & \textbf{1}, 2, 3                         \\
Batch Size               & 16, \textbf{32}                      \\
Learning Rate            & $3 \times 10^{-5}$\\
Temperature              & 0.01, \textbf{0.05}, 0.1   \\
Pooler                  & \textbf{cls}  \\ \bottomrule
\end{tabular}
\caption{Hyperparameters tested for training SimCSE with the unsupervised approach.}
\label{tab:hyperparams-unsupervised}
\end{table}

The time taken for the supervised SimCSE hyperparameter search is roughly 6-7 hours, and the time taken for the unsupervised SimCSE hyperparameter search is approximately 15-16 hours.

\subsection{Computing Infrastructure}

The following are the main libraries and the computing environment used in our experiments.

\begin{itemize}
    \item PyTorch: 1.13.0
    \item Huggingface's Transformers: 4.2.1
    \item SimCSE: 0.4
    \item NVIDIA A40 GPU
    \item CUDA 11.6
\end{itemize}

\subsection{Random Seeds}
In our experiments, we ensure that the same train-test splits are used across different runs of each experiment. Further, to control for any randomness throughout code execution, we
fixed the random seeds (to 42) in the following libraries:

\begin{itemize}
    \item Python (\texttt{random.seed});
    \item NumPy (\texttt{numpy.random.seed});
    \item PyTorch (\texttt{torch.manual\_seed});
    \item Tensorflow\\(\texttt{tensorflow.random.set\_seed}).
\end{itemize}

\subsection{Artifacts Used}
We primarily use two different types of artifacts, data and models. 

MFTC is a collection of 35,108 tweets annotated based on MFT \cite{Hoover2020}. MFTC can be accessed\footnote{\url{https://osf.io/k5n7y}} and used under Creative Commons Attribution 4.0 license.  MFD2.0 \cite{Frimer_2019} can be freely accessed\footnote{\url{https://osf.io/xakyw}}. 

SimCSE \cite{gao-etal-2021-simcse} can be used under MIT license\footnote{\url{https://github.com/princeton-nlp/SimCSE/blob/main/LICENSE}}. BERT \cite{devlin-etal-2019-bert} is used as a baseline model to compare with SimCSE. The license of BERT is Apache License 2.0\footnote{\url{https://github.com/google-research/bert/blob/master/LICENSE}}.

\renewcommand{\thefigure}{B\arabic{figure}}
\setcounter{figure}{0}
\renewcommand{\thetable}{B\arabic{table}}
\setcounter{table}{0}

\section{Extended Results}
We extend the results shown in the main paper for intrinsic and extrinsic evaluation.

\subsection{Intrinsic Evaluation}
We provide additional visualizations and quality metrics of the trained embedding spaces.

\subsubsection{Visualization}

\begin{figure}[!htb]
    \centering
    \includegraphics[width=0.9\linewidth]{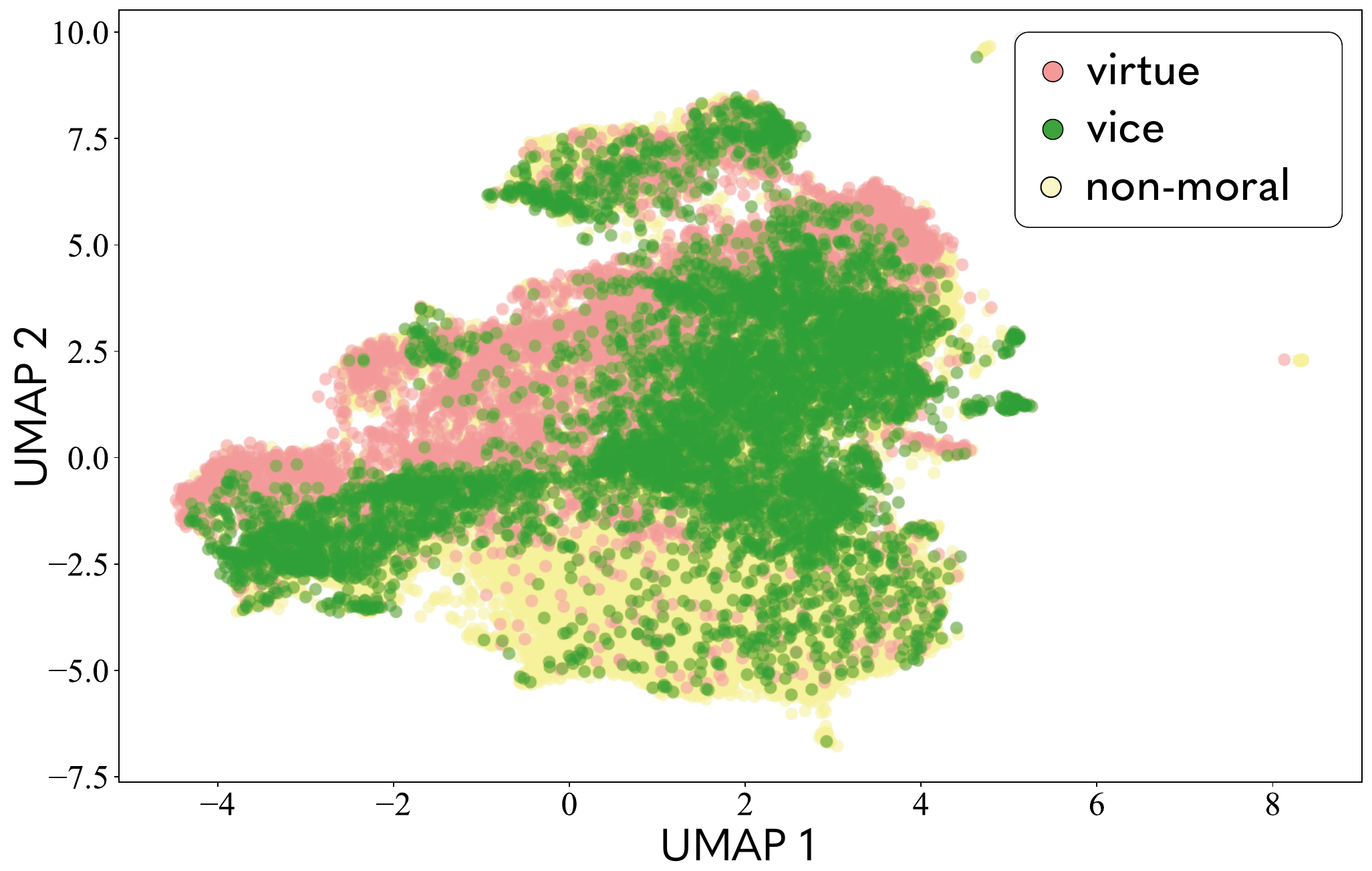}
    \caption{UMAP plot of MFTC training set with the off-the-shelf SimCSE model (only vices and virtues).}
    \label{fig:before_vice_virtue}
\end{figure}

\begin{figure}[!htb]
    \centering
    \includegraphics[width=0.9\linewidth]{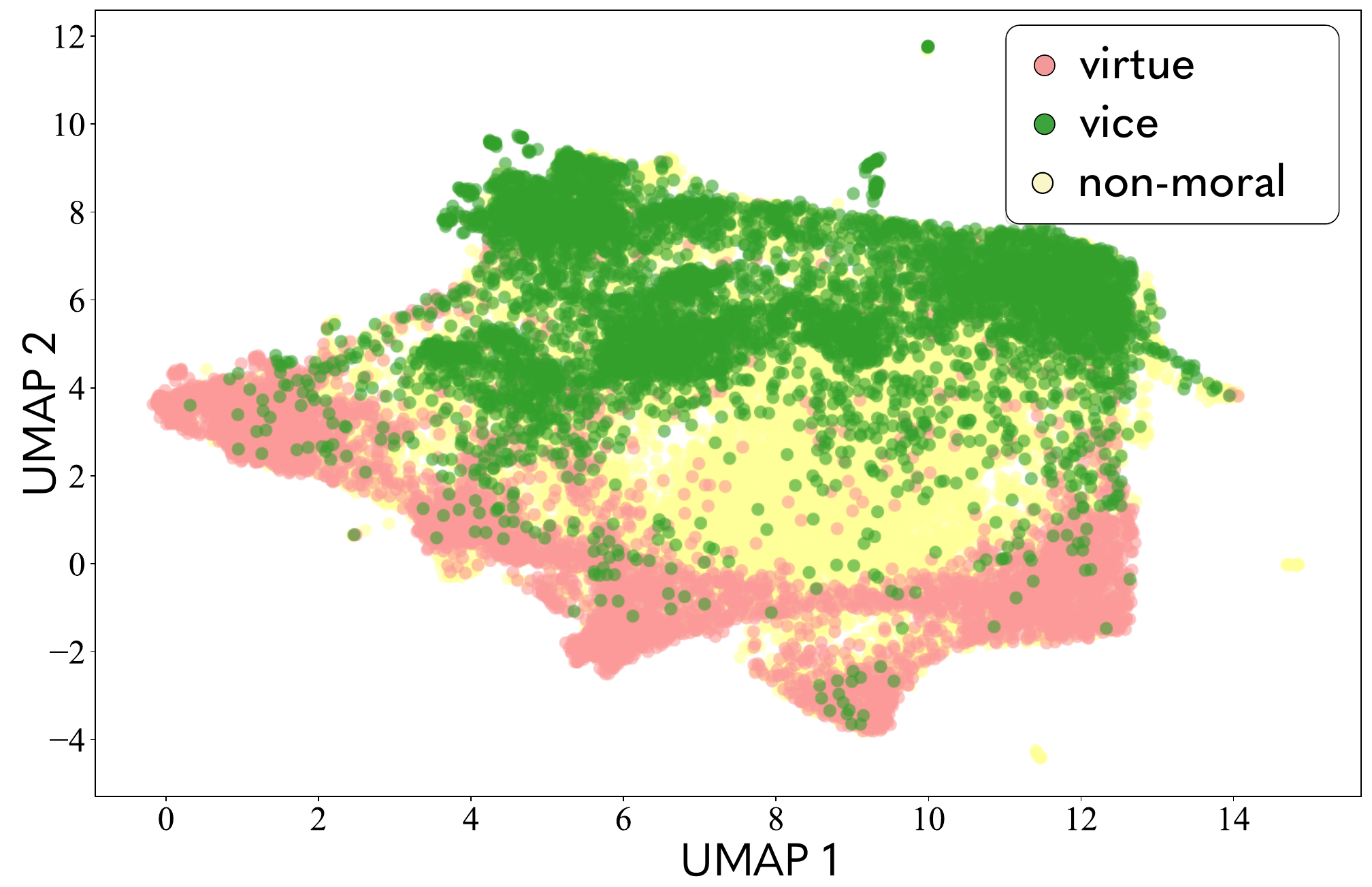}
    \caption{UMAP plot of MFTC training set with the supervised SimCSE approach (only vices and virtues).}
    \label{fig:umap_after_vice_virtue}
\end{figure}

Figures~\ref{fig:before_vice_virtue} and \ref{fig:umap_after_vice_virtue} show the UMAP plot of the MFTC training set mapped on the off-the-shelf SimcSE model the supervised SimCSE approach, respectively. The figures are similar to Figure~\ref{fig:umap}, however grouping the 10 moral elements as vices or virtues.

Figure~\ref{fig:before_vice_virtue} does not show any distinguishable cluster. Instead, Figure \ref{fig:umap_after_vice_virtue} shows a clearer separation between vice and virtue elements---vice and virtue clusters are less mixed together, and a bigger gap can be found between them.

\subsubsection{Moral Similarity}
\label{app:intrinsic-moral-sim}

\rebuttal{In the main paper we show the moral similarity table for the supervised SimCSE approach, here we show for the off-the-shelf model (Table~\ref{tab:moral-sim-sup-base-MFTC-train}) and for the unsupervised SimCSE approach (Table~\ref{tab:moral-sim-unsup-MFTC-train}).
Both tables show relatively low similarity along the diagonal when compared to Table~\ref{tab:moral-sim-MFTC}. The diagonal similarity of the virtue elements is higher than the vice elements for both tables, suggesting that a limited level of knowledge is already present in the off-the-shelf SimCSE. Moreover, the poor result of the unsupervised SimCSE approach aligns with the findings in the main paper, indicating that labels are necessary to grasp a pluralist approach to morality.}

\begin{table}[!htb]
\centering 
\hspace{-0.68cm}
\resizebox{0.98\linewidth}{!}{
\scriptsize
\renewcommand{\tabcolsep}{1.3pt}
\renewcommand{\arraystretch}{1.4} 
\begin{tabular}{rccccc|ccccc|c}
 & \rot{Care} & \rot{Fairness} & \rot{Loyalty} & \rot{Authority} & \rot{Purity} & \rot{Harm} & \rot{Cheating} & \rot{Betrayal} & \rot{Subversion} & \rot{Degradation} &
 \rot{Non-moral}\\
Care & {\cellcolor[HTML]{D5D5E8}} \color[HTML]{000000} 27.8 & {\cellcolor[HTML]{FEF6FA}} \color[HTML]{000000} 19.3 & {\cellcolor[HTML]{F4EDF6}} \color[HTML]{000000} 21.8 & {\cellcolor[HTML]{FFF7FB}} \color[HTML]{000000} 17.6 & {\cellcolor[HTML]{F9F2F8}} \color[HTML]{000000} 20.5 & {\cellcolor[HTML]{FFF7FB}} \color[HTML]{000000} 14.6 & {\cellcolor[HTML]{FFF7FB}} \color[HTML]{000000} 10.4 & {\cellcolor[HTML]{FFF7FB}} \color[HTML]{000000} 11.6 & {\cellcolor[HTML]{FFF7FB}} \color[HTML]{000000} 11.9 & {\cellcolor[HTML]{FFF7FB}} \color[HTML]{000000} 8.6 & {\cellcolor[HTML]{FFF7FB}} \color[HTML]{000000} 10.3 \\
Fairness & {\cellcolor[HTML]{FEF6FA}} \color[HTML]{000000} 19.3 & {\cellcolor[HTML]{C6CCE3}} \color[HTML]{000000} 29.7 & {\cellcolor[HTML]{EDE7F2}} \color[HTML]{000000} 23.7 & {\cellcolor[HTML]{F9F2F8}} \color[HTML]{000000} 20.5 & {\cellcolor[HTML]{FFF7FB}} \color[HTML]{000000} 18.2 & {\cellcolor[HTML]{FFF7FB}} \color[HTML]{000000} 16.9 & {\cellcolor[HTML]{FFF7FB}} \color[HTML]{000000} 17.5 & {\cellcolor[HTML]{FFF7FB}} \color[HTML]{000000} 17.2 & {\cellcolor[HTML]{FFF7FB}} \color[HTML]{000000} 17.1 & {\cellcolor[HTML]{FFF7FB}} \color[HTML]{000000} 12.6 & {\cellcolor[HTML]{FFF7FB}} \color[HTML]{000000} 11.6 \\
Loyalty & {\cellcolor[HTML]{F4EDF6}} \color[HTML]{000000} 21.8 & {\cellcolor[HTML]{EDE7F2}} \color[HTML]{000000} 23.7 & {\cellcolor[HTML]{D1D2E6}} \color[HTML]{000000} 28.5 & {\cellcolor[HTML]{FFF7FB}} \color[HTML]{000000} 18.4 & {\cellcolor[HTML]{FFF7FB}} \color[HTML]{000000} 17.5 & {\cellcolor[HTML]{FFF7FB}} \color[HTML]{000000} 14.7 & {\cellcolor[HTML]{FFF7FB}} \color[HTML]{000000} 13.8 & {\cellcolor[HTML]{FFF7FB}} \color[HTML]{000000} 16.8 & {\cellcolor[HTML]{FFF7FB}} \color[HTML]{000000} 16.0 & {\cellcolor[HTML]{FFF7FB}} \color[HTML]{000000} 9.9 & {\cellcolor[HTML]{FFF7FB}} \color[HTML]{000000} 11.4 \\
Authority & {\cellcolor[HTML]{FFF7FB}} \color[HTML]{000000} 17.6 & {\cellcolor[HTML]{F9F2F8}} \color[HTML]{000000} 20.5 & {\cellcolor[HTML]{FFF7FB}} \color[HTML]{000000} 18.4 & {\cellcolor[HTML]{F1EBF5}} \color[HTML]{000000} 22.5 & {\cellcolor[HTML]{FFF7FB}} \color[HTML]{000000} 16.4 & {\cellcolor[HTML]{FFF7FB}} \color[HTML]{000000} 13.0 & {\cellcolor[HTML]{FFF7FB}} \color[HTML]{000000} 13.7 & {\cellcolor[HTML]{FFF7FB}} \color[HTML]{000000} 14.6 & {\cellcolor[HTML]{FFF7FB}} \color[HTML]{000000} 15.7 & {\cellcolor[HTML]{FFF7FB}} \color[HTML]{000000} 10.4 & {\cellcolor[HTML]{FFF7FB}} \color[HTML]{000000} 10.2 \\
Purity & {\cellcolor[HTML]{F9F2F8}} \color[HTML]{000000} 20.5 & {\cellcolor[HTML]{FFF7FB}} \color[HTML]{000000} 18.2 & {\cellcolor[HTML]{FFF7FB}} \color[HTML]{000000} 17.5 & {\cellcolor[HTML]{FFF7FB}} \color[HTML]{000000} 16.4 & {\cellcolor[HTML]{E2DFEE}} \color[HTML]{000000} 25.5 & {\cellcolor[HTML]{FFF7FB}} \color[HTML]{000000} 10.9 & {\cellcolor[HTML]{FFF7FB}} \color[HTML]{000000} 9.8 & {\cellcolor[HTML]{FFF7FB}} \color[HTML]{000000} 10.2 & {\cellcolor[HTML]{FFF7FB}} \color[HTML]{000000} 10.3 & {\cellcolor[HTML]{FFF7FB}} \color[HTML]{000000} 9.8 & {\cellcolor[HTML]{FFF7FB}} \color[HTML]{000000} 8.7 \\
\hhline{~-----------}
Harm & {\cellcolor[HTML]{FFF7FB}} \color[HTML]{000000} 14.6 & {\cellcolor[HTML]{FFF7FB}} \color[HTML]{000000} 16.9 & {\cellcolor[HTML]{FFF7FB}} \color[HTML]{000000} 14.7 & {\cellcolor[HTML]{FFF7FB}} \color[HTML]{000000} 13.0 & {\cellcolor[HTML]{FFF7FB}} \color[HTML]{000000} 10.9 & {\cellcolor[HTML]{F3EDF5}} \color[HTML]{000000} 22.0 & {\cellcolor[HTML]{FFF7FB}} \color[HTML]{000000} 18.9 & {\cellcolor[HTML]{FDF5FA}} \color[HTML]{000000} 19.5 & {\cellcolor[HTML]{FFF7FB}} \color[HTML]{000000} 18.5 & {\cellcolor[HTML]{FFF7FB}} \color[HTML]{000000} 18.3 & {\cellcolor[HTML]{FFF7FB}} \color[HTML]{000000} 12.2 \\
Cheating & {\cellcolor[HTML]{FFF7FB}} \color[HTML]{000000} 10.4 & {\cellcolor[HTML]{FFF7FB}} \color[HTML]{000000} 17.5 & {\cellcolor[HTML]{FFF7FB}} \color[HTML]{000000} 13.8 & {\cellcolor[HTML]{FFF7FB}} \color[HTML]{000000} 13.7 & {\cellcolor[HTML]{FFF7FB}} \color[HTML]{000000} 9.8 & {\cellcolor[HTML]{FFF7FB}} \color[HTML]{000000} 18.9 & {\cellcolor[HTML]{F1EBF5}} \color[HTML]{000000} 22.4 & {\cellcolor[HTML]{F9F2F8}} \color[HTML]{000000} 20.5 & {\cellcolor[HTML]{FCF4FA}} \color[HTML]{000000} 19.6 & {\cellcolor[HTML]{FDF5FA}} \color[HTML]{000000} 19.5 & {\cellcolor[HTML]{FFF7FB}} \color[HTML]{000000} 11.9 \\
Betrayal & {\cellcolor[HTML]{FFF7FB}} \color[HTML]{000000} 11.6 & {\cellcolor[HTML]{FFF7FB}} \color[HTML]{000000} 17.2 & {\cellcolor[HTML]{FFF7FB}} \color[HTML]{000000} 16.8 & {\cellcolor[HTML]{FFF7FB}} \color[HTML]{000000} 14.6 & {\cellcolor[HTML]{FFF7FB}} \color[HTML]{000000} 10.2 & {\cellcolor[HTML]{FDF5FA}} \color[HTML]{000000} 19.5 & {\cellcolor[HTML]{F9F2F8}} \color[HTML]{000000} 20.5 & {\cellcolor[HTML]{EFE9F3}} \color[HTML]{000000} 23.0 & {\cellcolor[HTML]{F7F0F7}} \color[HTML]{000000} 20.9 & {\cellcolor[HTML]{FFF7FB}} \color[HTML]{000000} 18.4 & {\cellcolor[HTML]{FFF7FB}} \color[HTML]{000000} 12.3 \\
Subversion & {\cellcolor[HTML]{FFF7FB}} \color[HTML]{000000} 11.9 & {\cellcolor[HTML]{FFF7FB}} \color[HTML]{000000} 17.1 & {\cellcolor[HTML]{FFF7FB}} \color[HTML]{000000} 16.0 & {\cellcolor[HTML]{FFF7FB}} \color[HTML]{000000} 15.7 & {\cellcolor[HTML]{FFF7FB}} \color[HTML]{000000} 10.3 & {\cellcolor[HTML]{FFF7FB}} \color[HTML]{000000} 18.5 & {\cellcolor[HTML]{FCF4FA}} \color[HTML]{000000} 19.6 & {\cellcolor[HTML]{F7F0F7}} \color[HTML]{000000} 20.9 & {\cellcolor[HTML]{F3EDF5}} \color[HTML]{000000} 22.0 & {\cellcolor[HTML]{FFF7FB}} \color[HTML]{000000} 17.7 & {\cellcolor[HTML]{FFF7FB}} \color[HTML]{000000} 12.0 \\
Degradation & {\cellcolor[HTML]{FFF7FB}} \color[HTML]{000000} 8.6 & {\cellcolor[HTML]{FFF7FB}} \color[HTML]{000000} 12.6 & {\cellcolor[HTML]{FFF7FB}} \color[HTML]{000000} 9.9 & {\cellcolor[HTML]{FFF7FB}} \color[HTML]{000000} 10.4 & {\cellcolor[HTML]{FFF7FB}} \color[HTML]{000000} 9.8 & {\cellcolor[HTML]{FFF7FB}} \color[HTML]{000000} 18.3 & {\cellcolor[HTML]{FDF5FA}} \color[HTML]{000000} 19.5 & {\cellcolor[HTML]{FFF7FB}} \color[HTML]{000000} 18.4 & {\cellcolor[HTML]{FFF7FB}} \color[HTML]{000000} 17.7 & {\cellcolor[HTML]{EDE7F2}} \color[HTML]{000000} 23.7 & {\cellcolor[HTML]{FFF7FB}} \color[HTML]{000000} 11.9 \\
\hhline{~-----------}
Non-Moral & {\cellcolor[HTML]{FFF7FB}} \color[HTML]{000000} 10.3 & {\cellcolor[HTML]{FFF7FB}} \color[HTML]{000000} 11.6 & {\cellcolor[HTML]{FFF7FB}} \color[HTML]{000000} 11.4 & {\cellcolor[HTML]{FFF7FB}} \color[HTML]{000000} 10.2 & {\cellcolor[HTML]{FFF7FB}} \color[HTML]{000000} 8.7 & {\cellcolor[HTML]{FFF7FB}} \color[HTML]{000000} 12.2 & {\cellcolor[HTML]{FFF7FB}} \color[HTML]{000000} 11.9 & {\cellcolor[HTML]{FFF7FB}} \color[HTML]{000000} 12.3 & {\cellcolor[HTML]{FFF7FB}} \color[HTML]{000000} 12.0 & {\cellcolor[HTML]{FFF7FB}} \color[HTML]{000000} 11.9 & {\cellcolor[HTML]{FFF7FB}} \color[HTML]{000000} 9.8 \\

\end{tabular}}
\caption{Moral similarity on MFTC train set using the off-the-shelf SimCSE model.}
\label{tab:moral-sim-sup-base-MFTC-train}
\end{table}
\begin{table}[!htb]
\centering 
\hspace{-0.68cm}
\resizebox{0.98\linewidth}{!}{
\scriptsize
\renewcommand{\tabcolsep}{1.3pt}
\renewcommand{\arraystretch}{1.4} 
\begin{tabular}{rccccc|ccccc|c}
 & \rot{Care} & \rot{Fairness} & \rot{Loyalty} & \rot{Authority} & \rot{Purity} & \rot{Harm} & \rot{Cheating} & \rot{Betrayal} & \rot{Subversion} & \rot{Degradation} &
 \rot{Non-moral}\\
Care & {\cellcolor[HTML]{DFDDEC}} \color[HTML]{000000} 26.1 & {\cellcolor[HTML]{FDF5FA}} \color[HTML]{000000} 19.4 & {\cellcolor[HTML]{F4EEF6}} \color[HTML]{000000} 21.7 & {\cellcolor[HTML]{F5EEF6}} \color[HTML]{000000} 21.5 & {\cellcolor[HTML]{EEE9F3}} \color[HTML]{000000} 23.2 & {\cellcolor[HTML]{FCF4FA}} \color[HTML]{000000} 19.7 & {\cellcolor[HTML]{FFF7FB}} \color[HTML]{000000} 18.3 & {\cellcolor[HTML]{FFF7FB}} \color[HTML]{000000} 18.9 & {\cellcolor[HTML]{FEF6FA}} \color[HTML]{000000} 19.3 & {\cellcolor[HTML]{FFF7FB}} \color[HTML]{000000} 19.0 & {\cellcolor[HTML]{FDF5FA}} \color[HTML]{000000} 19.6 \\
Fairness & {\cellcolor[HTML]{FDF5FA}} \color[HTML]{000000} 19.4 & {\cellcolor[HTML]{E4E1EF}} \color[HTML]{000000} 25.1 & {\cellcolor[HTML]{F8F1F8}} \color[HTML]{000000} 20.8 & {\cellcolor[HTML]{F4EDF6}} \color[HTML]{000000} 21.8 & {\cellcolor[HTML]{FAF3F9}} \color[HTML]{000000} 20.2 & {\cellcolor[HTML]{FFF7FB}} \color[HTML]{000000} 18.7 & {\cellcolor[HTML]{FAF2F8}} \color[HTML]{000000} 20.4 & {\cellcolor[HTML]{FDF5FA}} \color[HTML]{000000} 19.6 & {\cellcolor[HTML]{FAF2F8}} \color[HTML]{000000} 20.3 & {\cellcolor[HTML]{FFF7FB}} \color[HTML]{000000} 18.6 & {\cellcolor[HTML]{FFF7FB}} \color[HTML]{000000} 19.0 \\
Loyalty & {\cellcolor[HTML]{F4EEF6}} \color[HTML]{000000} 21.7 & {\cellcolor[HTML]{F8F1F8}} \color[HTML]{000000} 20.8 & {\cellcolor[HTML]{E2DFEE}} \color[HTML]{000000} 25.5 & {\cellcolor[HTML]{F4EDF6}} \color[HTML]{000000} 21.8 & {\cellcolor[HTML]{F7F0F7}} \color[HTML]{000000} 21.1 & {\cellcolor[HTML]{FFF7FB}} \color[HTML]{000000} 18.8 & {\cellcolor[HTML]{FFF7FB}} \color[HTML]{000000} 18.8 & {\cellcolor[HTML]{F7F0F7}} \color[HTML]{000000} 20.9 & {\cellcolor[HTML]{F7F0F7}} \color[HTML]{000000} 21.0 & {\cellcolor[HTML]{FFF7FB}} \color[HTML]{000000} 18.7 & {\cellcolor[HTML]{FBF3F9}} \color[HTML]{000000} 20.0 \\
Authority & {\cellcolor[HTML]{F5EEF6}} \color[HTML]{000000} 21.5 & {\cellcolor[HTML]{F4EDF6}} \color[HTML]{000000} 21.8 & {\cellcolor[HTML]{F4EDF6}} \color[HTML]{000000} 21.8 & {\cellcolor[HTML]{DCDAEB}} \color[HTML]{000000} 26.6 & {\cellcolor[HTML]{F2ECF5}} \color[HTML]{000000} 22.3 & {\cellcolor[HTML]{FDF5FA}} \color[HTML]{000000} 19.6 & {\cellcolor[HTML]{F8F1F8}} \color[HTML]{000000} 20.8 & {\cellcolor[HTML]{F4EEF6}} \color[HTML]{000000} 21.7 & {\cellcolor[HTML]{EFE9F3}} \color[HTML]{000000} 23.1 & {\cellcolor[HTML]{FBF4F9}} \color[HTML]{000000} 19.9 & {\cellcolor[HTML]{F8F1F8}} \color[HTML]{000000} 20.7 \\
Purity & {\cellcolor[HTML]{EEE9F3}} \color[HTML]{000000} 23.2 & {\cellcolor[HTML]{FAF3F9}} \color[HTML]{000000} 20.2 & {\cellcolor[HTML]{F7F0F7}} \color[HTML]{000000} 21.1 & {\cellcolor[HTML]{F2ECF5}} \color[HTML]{000000} 22.3 & {\cellcolor[HTML]{D7D6E9}} \color[HTML]{000000} 27.5 & {\cellcolor[HTML]{FFF7FB}} \color[HTML]{000000} 18.4 & {\cellcolor[HTML]{FFF7FB}} \color[HTML]{000000} 18.7 & {\cellcolor[HTML]{FEF6FB}} \color[HTML]{000000} 19.1 & {\cellcolor[HTML]{FCF4FA}} \color[HTML]{000000} 19.7 & {\cellcolor[HTML]{FAF3F9}} \color[HTML]{000000} 20.1 & {\cellcolor[HTML]{FDF5FA}} \color[HTML]{000000} 19.4 \\
\hhline{~-----------}
Harm & {\cellcolor[HTML]{FCF4FA}} \color[HTML]{000000} 19.7 & {\cellcolor[HTML]{FFF7FB}} \color[HTML]{000000} 18.7 & {\cellcolor[HTML]{FFF7FB}} \color[HTML]{000000} 18.8 & {\cellcolor[HTML]{FDF5FA}} \color[HTML]{000000} 19.6 & {\cellcolor[HTML]{FFF7FB}} \color[HTML]{000000} 18.4 & {\cellcolor[HTML]{F2ECF5}} \color[HTML]{000000} 22.3 & {\cellcolor[HTML]{FAF2F8}} \color[HTML]{000000} 20.4 & {\cellcolor[HTML]{F8F1F8}} \color[HTML]{000000} 20.8 & {\cellcolor[HTML]{F7F0F7}} \color[HTML]{000000} 21.0 & {\cellcolor[HTML]{FAF2F8}} \color[HTML]{000000} 20.3 & {\cellcolor[HTML]{FDF5FA}} \color[HTML]{000000} 19.3 \\
Cheating & {\cellcolor[HTML]{FFF7FB}} \color[HTML]{000000} 18.3 & {\cellcolor[HTML]{FAF2F8}} \color[HTML]{000000} 20.4 & {\cellcolor[HTML]{FFF7FB}} \color[HTML]{000000} 18.8 & {\cellcolor[HTML]{F8F1F8}} \color[HTML]{000000} 20.8 & {\cellcolor[HTML]{FFF7FB}} \color[HTML]{000000} 18.7 & {\cellcolor[HTML]{FAF2F8}} \color[HTML]{000000} 20.4 & {\cellcolor[HTML]{EFE9F3}} \color[HTML]{000000} 23.1 & {\cellcolor[HTML]{F5EFF6}} \color[HTML]{000000} 21.4 & {\cellcolor[HTML]{F4EDF6}} \color[HTML]{000000} 21.9 & {\cellcolor[HTML]{F8F1F8}} \color[HTML]{000000} 20.8 & {\cellcolor[HTML]{FCF4FA}} \color[HTML]{000000} 19.7 \\
Betrayal & {\cellcolor[HTML]{FFF7FB}} \color[HTML]{000000} 18.9 & {\cellcolor[HTML]{FDF5FA}} \color[HTML]{000000} 19.6 & {\cellcolor[HTML]{F7F0F7}} \color[HTML]{000000} 20.9 & {\cellcolor[HTML]{F4EEF6}} \color[HTML]{000000} 21.7 & {\cellcolor[HTML]{FEF6FB}} \color[HTML]{000000} 19.1 & {\cellcolor[HTML]{F8F1F8}} \color[HTML]{000000} 20.8 & {\cellcolor[HTML]{F5EFF6}} \color[HTML]{000000} 21.4 & {\cellcolor[HTML]{EFE9F3}} \color[HTML]{000000} 23.1 & {\cellcolor[HTML]{F0EAF4}} \color[HTML]{000000} 22.9 & {\cellcolor[HTML]{F8F1F8}} \color[HTML]{000000} 20.6 & {\cellcolor[HTML]{FBF3F9}} \color[HTML]{000000} 20.0 \\
Subversion & {\cellcolor[HTML]{FEF6FA}} \color[HTML]{000000} 19.3 & {\cellcolor[HTML]{FAF2F8}} \color[HTML]{000000} 20.3 & {\cellcolor[HTML]{F7F0F7}} \color[HTML]{000000} 21.0 & {\cellcolor[HTML]{EFE9F3}} \color[HTML]{000000} 23.1 & {\cellcolor[HTML]{FCF4FA}} \color[HTML]{000000} 19.7 & {\cellcolor[HTML]{F7F0F7}} \color[HTML]{000000} 21.0 & {\cellcolor[HTML]{F4EDF6}} \color[HTML]{000000} 21.9 & {\cellcolor[HTML]{F0EAF4}} \color[HTML]{000000} 22.9 & {\cellcolor[HTML]{E9E5F1}} \color[HTML]{000000} 24.4 & {\cellcolor[HTML]{F7F0F7}} \color[HTML]{000000} 21.0 & {\cellcolor[HTML]{F9F2F8}} \color[HTML]{000000} 20.5 \\
Degradation & {\cellcolor[HTML]{FFF7FB}} \color[HTML]{000000} 19.0 & {\cellcolor[HTML]{FFF7FB}} \color[HTML]{000000} 18.6 & {\cellcolor[HTML]{FFF7FB}} \color[HTML]{000000} 18.7 & {\cellcolor[HTML]{FBF4F9}} \color[HTML]{000000} 19.9 & {\cellcolor[HTML]{FAF3F9}} \color[HTML]{000000} 20.1 & {\cellcolor[HTML]{FAF2F8}} \color[HTML]{000000} 20.3 & {\cellcolor[HTML]{F8F1F8}} \color[HTML]{000000} 20.8 & {\cellcolor[HTML]{F8F1F8}} \color[HTML]{000000} 20.6 & {\cellcolor[HTML]{F7F0F7}} \color[HTML]{000000} 21.0 & {\cellcolor[HTML]{F0EAF4}} \color[HTML]{000000} 22.8 & {\cellcolor[HTML]{FDF5FA}} \color[HTML]{000000} 19.6 \\
\hhline{~-----------}
Non-Moral & {\cellcolor[HTML]{FDF5FA}} \color[HTML]{000000} 19.6 & {\cellcolor[HTML]{FFF7FB}} \color[HTML]{000000} 19.0 & {\cellcolor[HTML]{FBF3F9}} \color[HTML]{000000} 20.0 & {\cellcolor[HTML]{F8F1F8}} \color[HTML]{000000} 20.7 & {\cellcolor[HTML]{FDF5FA}} \color[HTML]{000000} 19.4 & {\cellcolor[HTML]{FDF5FA}} \color[HTML]{000000} 19.3 & {\cellcolor[HTML]{FCF4FA}} \color[HTML]{000000} 19.7 & {\cellcolor[HTML]{FBF3F9}} \color[HTML]{000000} 20.0 & {\cellcolor[HTML]{F9F2F8}} \color[HTML]{000000} 20.5 & {\cellcolor[HTML]{FDF5FA}} \color[HTML]{000000} 19.6 & {\cellcolor[HTML]{FAF2F8}} \color[HTML]{000000} 20.4 \\

\end{tabular}}
\caption{Moral similarity on the MFTC train set using the unsupervised SimCSE approach.}
\label{tab:moral-sim-unsup-MFTC-train}
\end{table}

\subsubsection{Alignment and Uniformity}
\label{app:align-uniform}
\textit{Alignment} and \textit{unifomity} are metrics commonly used to assess the quality of an embedding space, measuring \textit{alignment} between positive pairs and \textit{uniformity} of the embedding space \cite{gao-etal-2021-simcse}. They can be calculated as follows:

\begin{equation*}
\small
\begin{aligned}
\mathcal{L}_{\text {align }}(f ; \alpha) & \triangleq \mathbb{E}_{(x, y) \sim p_{\text {pos }}}\left[\|f(x)-f(y)\|_2^\alpha\right]
\end{aligned}
\label{eq:alignment}
\end{equation*}

\begin{equation*}
\small
\begin{aligned}
\mathcal{L}_{\text {uniform }}(f ; t) & \triangleq \log \mathbb{E}_{ x, y} \overset{\text{i.i.d}}{\sim}  p_{\text {data }}\left[e^{-t\|f(x)-f(y)\|_2^2}\right]
\end{aligned}
\label{eq:uniformity}
\end{equation*}

Our goal is to generate the best possible embedding space mapping for this corpus---however, we only train on a relatively small and limited corpus, and thus we do not strive for a state-of-the-art \textit{alignment} and \textit{uniformity}.
Nevertheless, for completeness, we report the \textit{alignment} and \textit{uniformity} using the test dataset. Table~\ref{tab:align-uniform} displays the result of \textit{alignment} and \textit{uniformity} metrics. 
\rebuttal{The supervised SimCSE outperforms in \textit{alignment}, but gets a worse score in \textit{uniformity} when compared to the other two approaches. } 
This is consistent with the findings in the SimCSE paper \cite{gao-etal-2021-simcse} where the supervised SimCSE amends the \textit{alignment} and the unsupervised SimCSE effectively improves \textit{uniformity}.

\begin{table}[!htb]
\small
\centering
\begin{tabular}{@{}lcc@{}}
\toprule
\textbf{Approach} & \textbf{Alignment} & \textbf{Uniformity} \\ \midrule
\rebuttal{Off-the-shelf SimCSE} & 1.49      & -3.13      \\
Unsupervised SimCSE & 1.50      & -3.12      \\
Supervised SimCSE    & 0.77     & -2.27      \\
\bottomrule
\end{tabular}
\caption{\textit{Alignment} and \textit{uniformity} on MFTC test dataset. For both, lower numbers are better.}
\label{tab:align-uniform}
\end{table}

\subsection{Extrinsic Evaluation}

\rebuttal{We provide additional details on generalizability and comparison to MFD2.0 evaluation results,}
and offer further insight through a classification task.

\subsubsection{Generalizability on Test Set}
\label{appendix-generalizability}

\begin{figure}[!htb]
    \centering
    \includegraphics[width=0.9\linewidth]{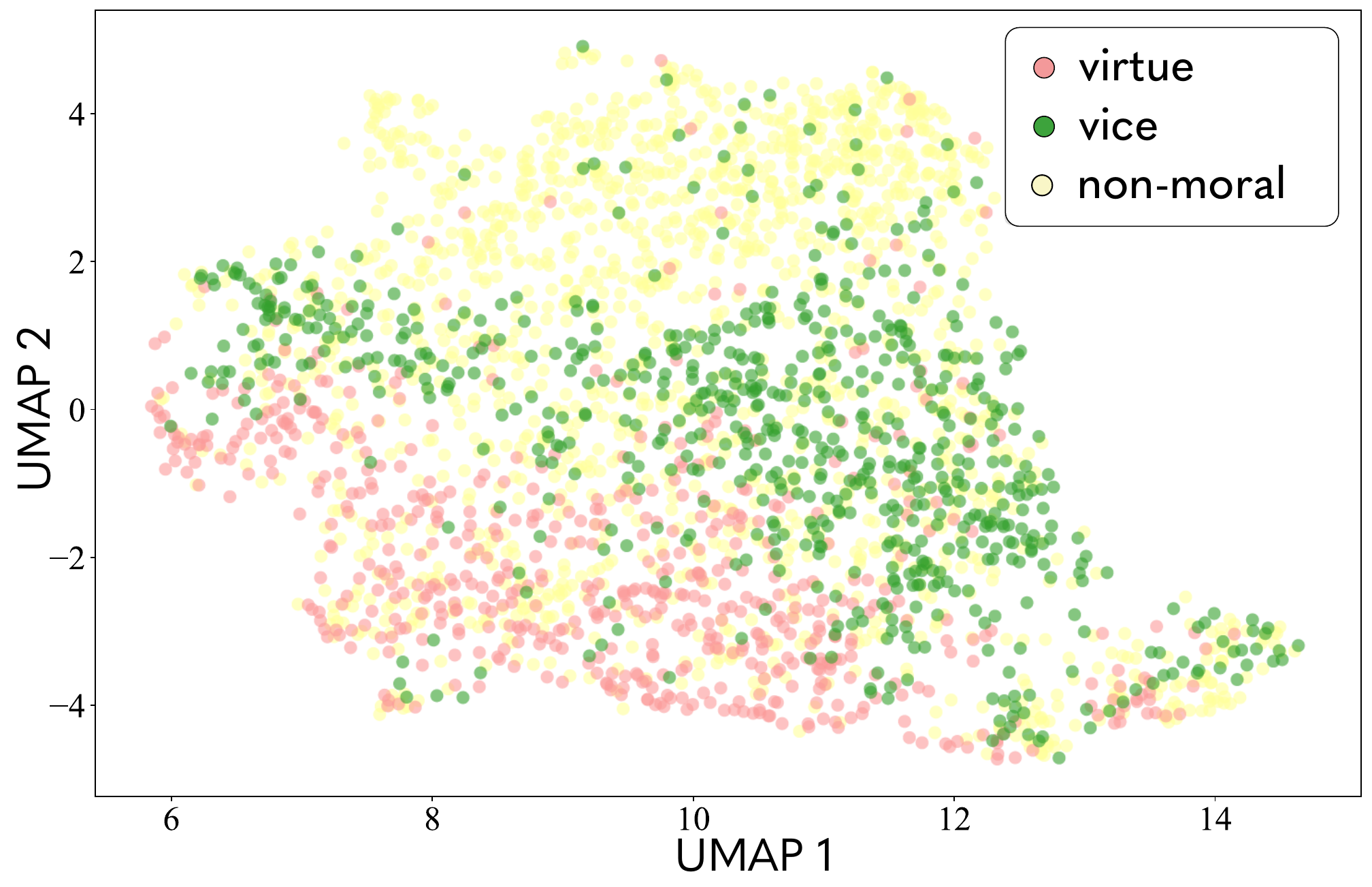}
    \caption{UMAP plot of MFTC test set with the off-the-shelf SimCSE model (only vices and virtues).}
    \label{fig:before_vice_virtue_test}
\end{figure}

\begin{figure}[!htb]
    \centering
    \includegraphics[width=0.9\linewidth]{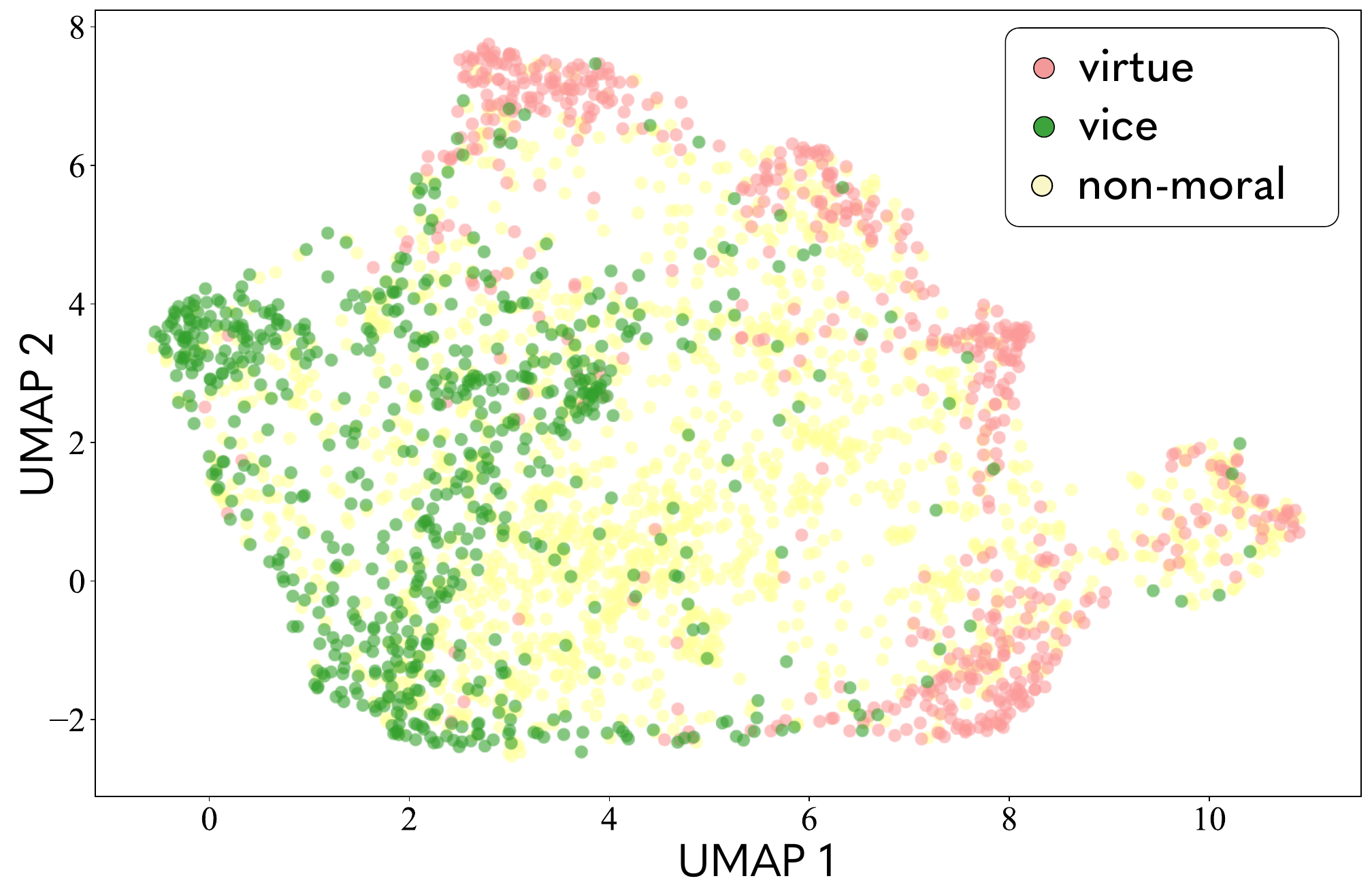}
    \caption{UMAP plot of MFTC test set with the supervies SimCSE approach (only vices and virtues).}
    \label{fig:umap_after_vice_virtue_test}
\end{figure}

Figures~\ref{fig:before_vice_virtue_test} and \ref{fig:umap_after_vice_virtue_test} show the UMAP plot of the MFTC test set mapped on the moral embedding space with the off-the-shelf model and with the supervised SimCSE approach, respectively. The figures are similar to Figure~\ref{fig:umap_moralD_after}, however grouping the 10 moral elements as vices or virtues. Figure~\ref{fig:before_vice_virtue_test} does not show clearly
distinguishable cluster. Instead, Figure \ref{fig:umap_after_vice_virtue_test} shows a clearer separation between vice and virtue values---vice and virtue clusters are less mixed together, and a bigger gap can be found between them.

\rebuttal{Table~\ref{tab:moral-sim-sup-base-MFTC-test} and Table~\ref{tab:moral-sim-unsup-MFTC-test} show the moral similarity obtained with off-the-shelf SimCSE model and unsupervised SimCSE approach (similar to Table~\ref{tab:moral-sim-MFTC-test}).
These tables confirm the visual intuition found in Figure~\ref{fig:umap_moralD_after}, with a low similarity along the diagonal. Further, these tables are consistent with the corresponding training set tables from the intrinsic evaluation (Tables~\ref{tab:moral-sim-sup-base-MFTC-train} and \ref{tab:moral-sim-unsup-MFTC-train}))}.

\begin{table}[!htb]
\centering 
\hspace{-0.68cm}
\resizebox{0.98\linewidth}{!}{
\scriptsize
\renewcommand{\tabcolsep}{1.3pt}
\renewcommand{\arraystretch}{1.4} 
\begin{tabular}{rccccc|ccccc|c}
 & \rot{Care} & \rot{Fairness} & \rot{Loyalty} & \rot{Authority} & \rot{Purity} & \rot{Harm} & \rot{Cheating} & \rot{Betrayal} & \rot{Subversion} & \rot{Degradation} &
 \rot{Non-moral}\\
Care & {\cellcolor[HTML]{D5D5E8}} \color[HTML]{000000} 27.8 & {\cellcolor[HTML]{FCF4FA}} \color[HTML]{000000} 19.7 & {\cellcolor[HTML]{F1EBF5}} \color[HTML]{000000} 22.4 & {\cellcolor[HTML]{FFF7FB}} \color[HTML]{000000} 17.5 & {\cellcolor[HTML]{F4EEF6}} \color[HTML]{000000} 21.8 & {\cellcolor[HTML]{FFF7FB}} \color[HTML]{000000} 13.9 & {\cellcolor[HTML]{FFF7FB}} \color[HTML]{000000} 10.6 & {\cellcolor[HTML]{FFF7FB}} \color[HTML]{000000} 10.8 & {\cellcolor[HTML]{FFF7FB}} \color[HTML]{000000} 10.8 & {\cellcolor[HTML]{FFF7FB}} \color[HTML]{000000} 7.8 & {\cellcolor[HTML]{FFF7FB}} \color[HTML]{000000} 10.4 \\
Fairness & {\cellcolor[HTML]{FCF4FA}} \color[HTML]{000000} 19.7 & {\cellcolor[HTML]{BFC9E1}} \color[HTML]{000000} 30.6 & {\cellcolor[HTML]{E9E5F1}} \color[HTML]{000000} 24.3 & {\cellcolor[HTML]{FAF3F9}} \color[HTML]{000000} 20.3 & {\cellcolor[HTML]{FAF3F9}} \color[HTML]{000000} 20.3 & {\cellcolor[HTML]{FFF7FB}} \color[HTML]{000000} 17.4 & {\cellcolor[HTML]{FFF7FB}} \color[HTML]{000000} 18.2 & {\cellcolor[HTML]{FFF7FB}} \color[HTML]{000000} 18.1 & {\cellcolor[HTML]{FFF7FB}} \color[HTML]{000000} 16.9 & {\cellcolor[HTML]{FFF7FB}} \color[HTML]{000000} 12.2 & {\cellcolor[HTML]{FFF7FB}} \color[HTML]{000000} 11.7 \\
Loyalty & {\cellcolor[HTML]{F1EBF5}} \color[HTML]{000000} 22.4 & {\cellcolor[HTML]{E9E5F1}} \color[HTML]{000000} 24.3 & {\cellcolor[HTML]{C9CEE4}} \color[HTML]{000000} 29.4 & {\cellcolor[HTML]{FFF7FB}} \color[HTML]{000000} 18.2 & {\cellcolor[HTML]{FFF7FB}} \color[HTML]{000000} 18.8 & {\cellcolor[HTML]{FFF7FB}} \color[HTML]{000000} 15.4 & {\cellcolor[HTML]{FFF7FB}} \color[HTML]{000000} 14.1 & {\cellcolor[HTML]{FFF7FB}} \color[HTML]{000000} 17.6 & {\cellcolor[HTML]{FFF7FB}} \color[HTML]{000000} 15.3 & {\cellcolor[HTML]{FFF7FB}} \color[HTML]{000000} 9.5 & {\cellcolor[HTML]{FFF7FB}} \color[HTML]{000000} 11.6 \\
Authority & {\cellcolor[HTML]{FFF7FB}} \color[HTML]{000000} 17.5 & {\cellcolor[HTML]{FAF3F9}} \color[HTML]{000000} 20.3 & {\cellcolor[HTML]{FFF7FB}} \color[HTML]{000000} 18.2 & {\cellcolor[HTML]{F2ECF5}} \color[HTML]{000000} 22.2 & {\cellcolor[HTML]{FFF7FB}} \color[HTML]{000000} 17.9 & {\cellcolor[HTML]{FFF7FB}} \color[HTML]{000000} 12.9 & {\cellcolor[HTML]{FFF7FB}} \color[HTML]{000000} 13.4 & {\cellcolor[HTML]{FFF7FB}} \color[HTML]{000000} 13.3 & {\cellcolor[HTML]{FFF7FB}} \color[HTML]{000000} 14.8 & {\cellcolor[HTML]{FFF7FB}} \color[HTML]{000000} 10.2 & {\cellcolor[HTML]{FFF7FB}} \color[HTML]{000000} 10.1 \\
Purity & {\cellcolor[HTML]{F4EEF6}} \color[HTML]{000000} 21.8 & {\cellcolor[HTML]{FAF3F9}} \color[HTML]{000000} 20.3 & {\cellcolor[HTML]{FFF7FB}} \color[HTML]{000000} 18.8 & {\cellcolor[HTML]{FFF7FB}} \color[HTML]{000000} 17.9 & {\cellcolor[HTML]{D1D2E6}} \color[HTML]{000000} 28.5 & {\cellcolor[HTML]{FFF7FB}} \color[HTML]{000000} 10.6 & {\cellcolor[HTML]{FFF7FB}} \color[HTML]{000000} 10.1 & {\cellcolor[HTML]{FFF7FB}} \color[HTML]{000000} 10.0 & {\cellcolor[HTML]{FFF7FB}} \color[HTML]{000000} 9.9 & {\cellcolor[HTML]{FFF7FB}} \color[HTML]{000000} 8.3 & {\cellcolor[HTML]{FFF7FB}} \color[HTML]{000000} 9.0 \\
\hhline{~-----------}
Harm & {\cellcolor[HTML]{FFF7FB}} \color[HTML]{000000} 13.9 & {\cellcolor[HTML]{FFF7FB}} \color[HTML]{000000} 17.4 & {\cellcolor[HTML]{FFF7FB}} \color[HTML]{000000} 15.4 & {\cellcolor[HTML]{FFF7FB}} \color[HTML]{000000} 12.9 & {\cellcolor[HTML]{FFF7FB}} \color[HTML]{000000} 10.6 & {\cellcolor[HTML]{F5EEF6}} \color[HTML]{000000} 21.5 & {\cellcolor[HTML]{FFF7FB}} \color[HTML]{000000} 18.4 & {\cellcolor[HTML]{FAF3F9}} \color[HTML]{000000} 20.1 & {\cellcolor[HTML]{FFF7FB}} \color[HTML]{000000} 17.9 & {\cellcolor[HTML]{FFF7FB}} \color[HTML]{000000} 17.0 & {\cellcolor[HTML]{FFF7FB}} \color[HTML]{000000} 11.8 \\
Cheating & {\cellcolor[HTML]{FFF7FB}} \color[HTML]{000000} 10.6 & {\cellcolor[HTML]{FFF7FB}} \color[HTML]{000000} 18.2 & {\cellcolor[HTML]{FFF7FB}} \color[HTML]{000000} 14.1 & {\cellcolor[HTML]{FFF7FB}} \color[HTML]{000000} 13.4 & {\cellcolor[HTML]{FFF7FB}} \color[HTML]{000000} 10.1 & {\cellcolor[HTML]{FFF7FB}} \color[HTML]{000000} 18.4 & {\cellcolor[HTML]{F0EAF4}} \color[HTML]{000000} 22.8 & {\cellcolor[HTML]{F5EEF6}} \color[HTML]{000000} 21.5 & {\cellcolor[HTML]{FFF7FB}} \color[HTML]{000000} 18.8 & {\cellcolor[HTML]{FFF7FB}} \color[HTML]{000000} 18.5 & {\cellcolor[HTML]{FFF7FB}} \color[HTML]{000000} 11.5 \\
Betrayal & {\cellcolor[HTML]{FFF7FB}} \color[HTML]{000000} 10.8 & {\cellcolor[HTML]{FFF7FB}} \color[HTML]{000000} 18.1 & {\cellcolor[HTML]{FFF7FB}} \color[HTML]{000000} 17.6 & {\cellcolor[HTML]{FFF7FB}} \color[HTML]{000000} 13.3 & {\cellcolor[HTML]{FFF7FB}} \color[HTML]{000000} 10.0 & {\cellcolor[HTML]{FAF3F9}} \color[HTML]{000000} 20.1 & {\cellcolor[HTML]{F5EEF6}} \color[HTML]{000000} 21.5 & {\cellcolor[HTML]{DEDCEC}} \color[HTML]{000000} 26.3 & {\cellcolor[HTML]{F7F0F7}} \color[HTML]{000000} 21.1 & {\cellcolor[HTML]{FDF5FA}} \color[HTML]{000000} 19.6 & {\cellcolor[HTML]{FFF7FB}} \color[HTML]{000000} 12.6 \\
Subversion & {\cellcolor[HTML]{FFF7FB}} \color[HTML]{000000} 10.8 & {\cellcolor[HTML]{FFF7FB}} \color[HTML]{000000} 16.9 & {\cellcolor[HTML]{FFF7FB}} \color[HTML]{000000} 15.3 & {\cellcolor[HTML]{FFF7FB}} \color[HTML]{000000} 14.8 & {\cellcolor[HTML]{FFF7FB}} \color[HTML]{000000} 9.9 & {\cellcolor[HTML]{FFF7FB}} \color[HTML]{000000} 17.9 & {\cellcolor[HTML]{FFF7FB}} \color[HTML]{000000} 18.8 & {\cellcolor[HTML]{F7F0F7}} \color[HTML]{000000} 21.1 & {\cellcolor[HTML]{F4EDF6}} \color[HTML]{000000} 21.8 & {\cellcolor[HTML]{FFF7FB}} \color[HTML]{000000} 17.6 & {\cellcolor[HTML]{FFF7FB}} \color[HTML]{000000} 11.3 \\
Degradation & {\cellcolor[HTML]{FFF7FB}} \color[HTML]{000000} 7.8 & {\cellcolor[HTML]{FFF7FB}} \color[HTML]{000000} 12.2 & {\cellcolor[HTML]{FFF7FB}} \color[HTML]{000000} 9.5 & {\cellcolor[HTML]{FFF7FB}} \color[HTML]{000000} 10.2 & {\cellcolor[HTML]{FFF7FB}} \color[HTML]{000000} 8.3 & {\cellcolor[HTML]{FFF7FB}} \color[HTML]{000000} 17.0 & {\cellcolor[HTML]{FFF7FB}} \color[HTML]{000000} 18.5 & {\cellcolor[HTML]{FDF5FA}} \color[HTML]{000000} 19.6 & {\cellcolor[HTML]{FFF7FB}} \color[HTML]{000000} 17.6 & {\cellcolor[HTML]{ECE7F2}} \color[HTML]{000000} 23.8 & {\cellcolor[HTML]{FFF7FB}} \color[HTML]{000000} 11.8 \\
\hhline{~-----------}
Non-Moral & {\cellcolor[HTML]{FFF7FB}} \color[HTML]{000000} 10.4 & {\cellcolor[HTML]{FFF7FB}} \color[HTML]{000000} 11.7 & {\cellcolor[HTML]{FFF7FB}} \color[HTML]{000000} 11.6 & {\cellcolor[HTML]{FFF7FB}} \color[HTML]{000000} 10.1 & {\cellcolor[HTML]{FFF7FB}} \color[HTML]{000000} 9.0 & {\cellcolor[HTML]{FFF7FB}} \color[HTML]{000000} 11.8 & {\cellcolor[HTML]{FFF7FB}} \color[HTML]{000000} 11.5 & {\cellcolor[HTML]{FFF7FB}} \color[HTML]{000000} 12.6 & {\cellcolor[HTML]{FFF7FB}} \color[HTML]{000000} 11.3 & {\cellcolor[HTML]{FFF7FB}} \color[HTML]{000000} 11.8 & {\cellcolor[HTML]{FFF7FB}} \color[HTML]{000000} 9.6 \\

\end{tabular}}
\caption{Moral similarity on MFTC test set using the off-the-shelf SimCSE model.}
\label{tab:moral-sim-sup-base-MFTC-test}
\end{table}
\begin{table}[!htb]
\centering 
\hspace{-0.68cm}
\resizebox{0.98\linewidth}{!}{
\scriptsize
\renewcommand{\tabcolsep}{1.3pt}
\renewcommand{\arraystretch}{1.4} 
\begin{tabular}{rccccc|ccccc|c}
 & \rot{Care} & \rot{Fairness} & \rot{Loyalty} & \rot{Authority} & \rot{Purity} & \rot{Harm} & \rot{Cheating} & \rot{Betrayal} & \rot{Subversion} & \rot{Degradation} &
 \rot{Non-moral}\\
Care & {\cellcolor[HTML]{D9D8EA}} \color[HTML]{000000} 27.1 & {\cellcolor[HTML]{FDF5FA}} \color[HTML]{000000} 19.5 & {\cellcolor[HTML]{F2ECF5}} \color[HTML]{000000} 22.1 & {\cellcolor[HTML]{F5EEF6}} \color[HTML]{000000} 21.6 & {\cellcolor[HTML]{EDE7F2}} \color[HTML]{000000} 23.6 & {\cellcolor[HTML]{FDF5FA}} \color[HTML]{000000} 19.4 & {\cellcolor[HTML]{FFF7FB}} \color[HTML]{000000} 18.3 & {\cellcolor[HTML]{FFF7FB}} \color[HTML]{000000} 18.3 & {\cellcolor[HTML]{FEF6FB}} \color[HTML]{000000} 19.1 & {\cellcolor[HTML]{FFF7FB}} \color[HTML]{000000} 18.6 & {\cellcolor[HTML]{FCF4FA}} \color[HTML]{000000} 19.7 \\
Fairness & {\cellcolor[HTML]{FDF5FA}} \color[HTML]{000000} 19.5 & {\cellcolor[HTML]{E4E1EF}} \color[HTML]{000000} 25.2 & {\cellcolor[HTML]{F7F0F7}} \color[HTML]{000000} 21.0 & {\cellcolor[HTML]{F3EDF5}} \color[HTML]{000000} 22.0 & {\cellcolor[HTML]{F7F0F7}} \color[HTML]{000000} 21.2 & {\cellcolor[HTML]{FFF7FB}} \color[HTML]{000000} 18.8 & {\cellcolor[HTML]{FAF2F8}} \color[HTML]{000000} 20.3 & {\cellcolor[HTML]{FEF6FB}} \color[HTML]{000000} 19.2 & {\cellcolor[HTML]{FAF2F8}} \color[HTML]{000000} 20.4 & {\cellcolor[HTML]{FEF6FB}} \color[HTML]{000000} 19.1 & {\cellcolor[HTML]{FFF7FB}} \color[HTML]{000000} 19.0 \\
Loyalty & {\cellcolor[HTML]{F2ECF5}} \color[HTML]{000000} 22.1 & {\cellcolor[HTML]{F7F0F7}} \color[HTML]{000000} 21.0 & {\cellcolor[HTML]{E0DDED}} \color[HTML]{000000} 26.0 & {\cellcolor[HTML]{F4EDF6}} \color[HTML]{000000} 21.8 & {\cellcolor[HTML]{F6EFF7}} \color[HTML]{000000} 21.3 & {\cellcolor[HTML]{FEF6FB}} \color[HTML]{000000} 19.2 & {\cellcolor[HTML]{FFF7FB}} \color[HTML]{000000} 18.6 & {\cellcolor[HTML]{F8F1F8}} \color[HTML]{000000} 20.6 & {\cellcolor[HTML]{F6EFF7}} \color[HTML]{000000} 21.2 & {\cellcolor[HTML]{FDF5FA}} \color[HTML]{000000} 19.6 & {\cellcolor[HTML]{FAF3F9}} \color[HTML]{000000} 20.1 \\
Authority & {\cellcolor[HTML]{F5EEF6}} \color[HTML]{000000} 21.6 & {\cellcolor[HTML]{F3EDF5}} \color[HTML]{000000} 22.0 & {\cellcolor[HTML]{F4EDF6}} \color[HTML]{000000} 21.8 & {\cellcolor[HTML]{D9D8EA}} \color[HTML]{000000} 27.0 & {\cellcolor[HTML]{F0EAF4}} \color[HTML]{000000} 22.8 & {\cellcolor[HTML]{FCF4FA}} \color[HTML]{000000} 19.7 & {\cellcolor[HTML]{F8F1F8}} \color[HTML]{000000} 20.7 & {\cellcolor[HTML]{F7F0F7}} \color[HTML]{000000} 21.0 & {\cellcolor[HTML]{F0EAF4}} \color[HTML]{000000} 23.0 & {\cellcolor[HTML]{F8F1F8}} \color[HTML]{000000} 20.7 & {\cellcolor[HTML]{F8F1F8}} \color[HTML]{000000} 20.8 \\
Purity & {\cellcolor[HTML]{EDE7F2}} \color[HTML]{000000} 23.6 & {\cellcolor[HTML]{F7F0F7}} \color[HTML]{000000} 21.2 & {\cellcolor[HTML]{F6EFF7}} \color[HTML]{000000} 21.3 & {\cellcolor[HTML]{F0EAF4}} \color[HTML]{000000} 22.8 & {\cellcolor[HTML]{CCCFE5}} \color[HTML]{000000} 29.2 & {\cellcolor[HTML]{FFF7FB}} \color[HTML]{000000} 18.4 & {\cellcolor[HTML]{FEF6FA}} \color[HTML]{000000} 19.2 & {\cellcolor[HTML]{FDF5FA}} \color[HTML]{000000} 19.5 & {\cellcolor[HTML]{FAF3F9}} \color[HTML]{000000} 20.1 & {\cellcolor[HTML]{FBF4F9}} \color[HTML]{000000} 19.9 & {\cellcolor[HTML]{FDF5FA}} \color[HTML]{000000} 19.6 \\
\hhline{~-----------}
Harm & {\cellcolor[HTML]{FDF5FA}} \color[HTML]{000000} 19.4 & {\cellcolor[HTML]{FFF7FB}} \color[HTML]{000000} 18.8 & {\cellcolor[HTML]{FEF6FB}} \color[HTML]{000000} 19.2 & {\cellcolor[HTML]{FCF4FA}} \color[HTML]{000000} 19.7 & {\cellcolor[HTML]{FFF7FB}} \color[HTML]{000000} 18.4 & {\cellcolor[HTML]{F1EBF4}} \color[HTML]{000000} 22.5 & {\cellcolor[HTML]{FAF2F8}} \color[HTML]{000000} 20.4 & {\cellcolor[HTML]{F8F1F8}} \color[HTML]{000000} 20.7 & {\cellcolor[HTML]{F5EFF6}} \color[HTML]{000000} 21.3 & {\cellcolor[HTML]{FAF2F8}} \color[HTML]{000000} 20.3 & {\cellcolor[HTML]{FDF5FA}} \color[HTML]{000000} 19.5 \\
Cheating & {\cellcolor[HTML]{FFF7FB}} \color[HTML]{000000} 18.3 & {\cellcolor[HTML]{FAF2F8}} \color[HTML]{000000} 20.3 & {\cellcolor[HTML]{FFF7FB}} \color[HTML]{000000} 18.6 & {\cellcolor[HTML]{F8F1F8}} \color[HTML]{000000} 20.7 & {\cellcolor[HTML]{FEF6FA}} \color[HTML]{000000} 19.2 & {\cellcolor[HTML]{FAF2F8}} \color[HTML]{000000} 20.4 & {\cellcolor[HTML]{EDE7F2}} \color[HTML]{000000} 23.7 & {\cellcolor[HTML]{F6EFF7}} \color[HTML]{000000} 21.3 & {\cellcolor[HTML]{F5EEF6}} \color[HTML]{000000} 21.6 & {\cellcolor[HTML]{F7F0F7}} \color[HTML]{000000} 20.9 & {\cellcolor[HTML]{FDF5FA}} \color[HTML]{000000} 19.6 \\
Betrayal & {\cellcolor[HTML]{FFF7FB}} \color[HTML]{000000} 18.3 & {\cellcolor[HTML]{FEF6FB}} \color[HTML]{000000} 19.2 & {\cellcolor[HTML]{F8F1F8}} \color[HTML]{000000} 20.6 & {\cellcolor[HTML]{F7F0F7}} \color[HTML]{000000} 21.0 & {\cellcolor[HTML]{FDF5FA}} \color[HTML]{000000} 19.5 & {\cellcolor[HTML]{F8F1F8}} \color[HTML]{000000} 20.7 & {\cellcolor[HTML]{F6EFF7}} \color[HTML]{000000} 21.3 & {\cellcolor[HTML]{E9E5F1}} \color[HTML]{000000} 24.2 & {\cellcolor[HTML]{F0EAF4}} \color[HTML]{000000} 22.8 & {\cellcolor[HTML]{F5EFF6}} \color[HTML]{000000} 21.4 & {\cellcolor[HTML]{FBF4F9}} \color[HTML]{000000} 19.9 \\
Subversion & {\cellcolor[HTML]{FEF6FB}} \color[HTML]{000000} 19.1 & {\cellcolor[HTML]{FAF2F8}} \color[HTML]{000000} 20.4 & {\cellcolor[HTML]{F6EFF7}} \color[HTML]{000000} 21.2 & {\cellcolor[HTML]{F0EAF4}} \color[HTML]{000000} 23.0 & {\cellcolor[HTML]{FAF3F9}} \color[HTML]{000000} 20.1 & {\cellcolor[HTML]{F5EFF6}} \color[HTML]{000000} 21.3 & {\cellcolor[HTML]{F5EEF6}} \color[HTML]{000000} 21.6 & {\cellcolor[HTML]{F0EAF4}} \color[HTML]{000000} 22.8 & {\cellcolor[HTML]{E6E2EF}} \color[HTML]{000000} 24.9 & {\cellcolor[HTML]{F4EDF6}} \color[HTML]{000000} 21.9 & {\cellcolor[HTML]{F8F1F8}} \color[HTML]{000000} 20.7 \\
Degradation & {\cellcolor[HTML]{FFF7FB}} \color[HTML]{000000} 18.6 & {\cellcolor[HTML]{FEF6FB}} \color[HTML]{000000} 19.1 & {\cellcolor[HTML]{FDF5FA}} \color[HTML]{000000} 19.6 & {\cellcolor[HTML]{F8F1F8}} \color[HTML]{000000} 20.7 & {\cellcolor[HTML]{FBF4F9}} \color[HTML]{000000} 19.9 & {\cellcolor[HTML]{FAF2F8}} \color[HTML]{000000} 20.3 & {\cellcolor[HTML]{F7F0F7}} \color[HTML]{000000} 20.9 & {\cellcolor[HTML]{F5EFF6}} \color[HTML]{000000} 21.4 & {\cellcolor[HTML]{F4EDF6}} \color[HTML]{000000} 21.9 & {\cellcolor[HTML]{EDE8F3}} \color[HTML]{000000} 23.6 & {\cellcolor[HTML]{FAF3F9}} \color[HTML]{000000} 20.2 \\
\hhline{~-----------}
Non-Moral & {\cellcolor[HTML]{FCF4FA}} \color[HTML]{000000} 19.7 & {\cellcolor[HTML]{FFF7FB}} \color[HTML]{000000} 19.0 & {\cellcolor[HTML]{FAF3F9}} \color[HTML]{000000} 20.1 & {\cellcolor[HTML]{F8F1F8}} \color[HTML]{000000} 20.8 & {\cellcolor[HTML]{FDF5FA}} \color[HTML]{000000} 19.6 & {\cellcolor[HTML]{FDF5FA}} \color[HTML]{000000} 19.5 & {\cellcolor[HTML]{FDF5FA}} \color[HTML]{000000} 19.6 & {\cellcolor[HTML]{FBF4F9}} \color[HTML]{000000} 19.9 & {\cellcolor[HTML]{F8F1F8}} \color[HTML]{000000} 20.7 & {\cellcolor[HTML]{FAF3F9}} \color[HTML]{000000} 20.2 & {\cellcolor[HTML]{F8F1F8}} \color[HTML]{000000} 20.6 \\

\end{tabular}}
\caption{Moral similarity on MFTC test set using the unsupervised SimCSE approach.}
\label{tab:moral-sim-unsup-MFTC-test}
\end{table}

\subsubsection{Comparison to MFD2.0}
\label{appendix:MFD2.0}

\paragraph{Clustering}

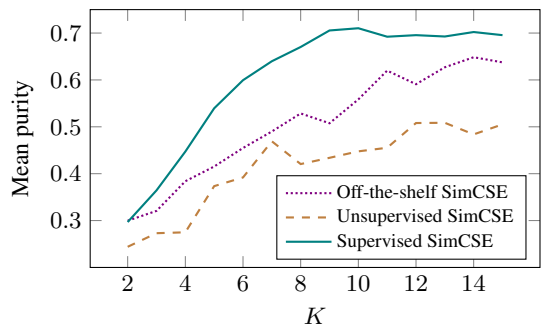
\begin{figure}[!htb]
\small
\begin{tikzpicture}
    \begin{axis}[
        width=7.5cm,
        height=5cm,
        xlabel = {$K$},
        ylabel = {Mean purity},
        ymin = 0.2, ymax = 0.75,
        xtick={2,4,6,8,10,12,14},
        ytick={0.3,0.4,0.5,0.6,0.7},
        legend style={nodes={scale=0.8}},
        legend cell align={left},
        legend pos = south east
    ]
    \addplot [violet, thick, densely dotted] table {tikz/purity/sup_base.dat};
    \addlegendentry{\rebuttal{Off-the-shelf SimCSE}};
    \addplot [brown, thick, dashed] table {tikz/purity/unsup.dat};
    \addlegendentry{Unsupervised SimCSE};
    \addplot [teal, thick] table {tikz/purity/sup.dat};
    \addlegendentry{Supervised SimCSE};
    \end{axis}
\end{tikzpicture}
\caption{Mean purity for $K$ ranging from 2 to 15 for the three compared embedding spaces.}
\label{fig:clustering-purity}
\end{figure}

In Figure~\ref{fig:clustering-purity} we report the purity score for $K$ ranging from 2 to 15 (similar to the Silhouette coefficient in Section~\ref{sec:result-clustering}).

We observe an overall increase in the mean purity score for all approaches as $K$ increases, which is to be expected due to the calculation of the purity score (Section~\ref{sec:clustering}).
We notice that the supervised SimCSE results in higher mean purity compared to other approaches, reaching its peak at $K=9$ and $K=10$. These values are similar to the number of moral values, indicating that corresponding embedding spaces are consistent with the MFT taxonomy and the MFD2.0 lexicon.
Further, we observe that the supervised SimCSE approach and the off-the-shelf SimCSE model lead to a higher mean purity compared to the unsupervised SimCSE approach.

\paragraph{Moral Similarity}

In Table~\ref{tab:moral-sim-mfd2} we report the moral similarity for MFD2.0 with the supervised SimCSE approach, whereas in Tables~\ref{tab:moral-sim-sup-base-mfd2} \rebuttal{and \ref{tab:moral-sim-mfd2-unsup}} we report the analogous results with the off-the-shelf model and the unsupervised SimCSE approach. We notice how the unsupervised approach only slightly captures the similarity among words belonging to the same MFT element, in strong contrast with the supervised approach.
\rebuttal{We observe the same pattern with off-the-shelf SimCSE approach in Table~\ref{tab:moral-sim-sup-base-mfd2}. The strong similarity of Tables~\ref{tab:moral-sim-sup-base-mfd2} and \ref{tab:moral-sim-mfd2-unsup} corresponds with the clustering findings described in Figure~\ref{fig:clustering-k} and Figure~\ref{fig:clustering-purity}, with the off-the-shelf SimCSE model leading to slightly better results to the unsupervised SimCSE approach.}

\begin{table}[!htb]
\centering
\scriptsize
\hspace{-0.7cm}
\renewcommand{\tabcolsep}{1.5pt}
\renewcommand{\arraystretch}{1.4} 
\begin{tabular}{rccccc|ccccc}
 & \rot{Care} & \rot{Fairness} & \rot{Loyalty} & \rot{Authority} & \rot{Purity} & \rot{Harm} & \rot{Cheating} & \rot{Betrayal} & \rot{Subversion} & \rot{Degradation} \\
Care & {\cellcolor[HTML]{B3C3DE}} \color[HTML]{000000} 32.0 & {\cellcolor[HTML]{FFF7FB}} \color[HTML]{000000} 18.0 & {\cellcolor[HTML]{FBF3F9}} \color[HTML]{000000} 20.0 & {\cellcolor[HTML]{FFF7FB}} \color[HTML]{000000} 17.1 & {\cellcolor[HTML]{FEF6FB}} \color[HTML]{000000} 19.2 & {\cellcolor[HTML]{FFF7FB}} \color[HTML]{000000} 16.6 & {\cellcolor[HTML]{FFF7FB}} \color[HTML]{000000} 13.3 & {\cellcolor[HTML]{FFF7FB}} \color[HTML]{000000} 13.3 & {\cellcolor[HTML]{FFF7FB}} \color[HTML]{000000} 12.3 & {\cellcolor[HTML]{FFF7FB}} \color[HTML]{000000} 13.6 \\
Fairness & {\cellcolor[HTML]{FFF7FB}} \color[HTML]{000000} 18.0 & {\cellcolor[HTML]{D3D4E7}} \color[HTML]{000000} 28.0 & {\cellcolor[HTML]{FFF7FB}} \color[HTML]{000000} 17.4 & {\cellcolor[HTML]{FFF7FB}} \color[HTML]{000000} 17.8 & {\cellcolor[HTML]{FFF7FB}} \color[HTML]{000000} 16.0 & {\cellcolor[HTML]{FFF7FB}} \color[HTML]{000000} 13.1 & {\cellcolor[HTML]{FFF7FB}} \color[HTML]{000000} 16.1 & {\cellcolor[HTML]{FFF7FB}} \color[HTML]{000000} 16.1 & {\cellcolor[HTML]{FFF7FB}} \color[HTML]{000000} 16.2 & {\cellcolor[HTML]{FFF7FB}} \color[HTML]{000000} 11.4 \\
Loyalty & {\cellcolor[HTML]{FBF3F9}} \color[HTML]{000000} 20.0 & {\cellcolor[HTML]{FFF7FB}} \color[HTML]{000000} 17.4 & {\cellcolor[HTML]{C4CBE3}} \color[HTML]{000000} 30.0 & {\cellcolor[HTML]{FAF3F9}} \color[HTML]{000000} 20.2 & {\cellcolor[HTML]{FFF7FB}} \color[HTML]{000000} 18.2 & {\cellcolor[HTML]{FFF7FB}} \color[HTML]{000000} 15.0 & {\cellcolor[HTML]{FFF7FB}} \color[HTML]{000000} 16.2 & {\cellcolor[HTML]{FAF2F8}} \color[HTML]{000000} 20.3 & {\cellcolor[HTML]{FBF4F9}} \color[HTML]{000000} 19.9 & {\cellcolor[HTML]{FFF7FB}} \color[HTML]{000000} 14.3 \\
Authority & {\cellcolor[HTML]{FFF7FB}} \color[HTML]{000000} 17.1 & {\cellcolor[HTML]{FFF7FB}} \color[HTML]{000000} 17.8 & {\cellcolor[HTML]{FAF3F9}} \color[HTML]{000000} 20.2 & {\cellcolor[HTML]{E2DFEE}} \color[HTML]{000000} 25.4 & {\cellcolor[HTML]{FFF7FB}} \color[HTML]{000000} 18.2 & {\cellcolor[HTML]{FFF7FB}} \color[HTML]{000000} 15.0 & {\cellcolor[HTML]{FFF7FB}} \color[HTML]{000000} 14.5 & {\cellcolor[HTML]{FFF7FB}} \color[HTML]{000000} 17.4 & {\cellcolor[HTML]{FFF7FB}} \color[HTML]{000000} 19.0 & {\cellcolor[HTML]{FFF7FB}} \color[HTML]{000000} 13.2 \\
Purity & {\cellcolor[HTML]{FEF6FB}} \color[HTML]{000000} 19.2 & {\cellcolor[HTML]{FFF7FB}} \color[HTML]{000000} 16.0 & {\cellcolor[HTML]{FFF7FB}} \color[HTML]{000000} 18.2 & {\cellcolor[HTML]{FFF7FB}} \color[HTML]{000000} 18.2 & {\cellcolor[HTML]{DEDCEC}} \color[HTML]{000000} 26.2 & {\cellcolor[HTML]{FFF7FB}} \color[HTML]{000000} 12.7 & {\cellcolor[HTML]{FFF7FB}} \color[HTML]{000000} 11.0 & {\cellcolor[HTML]{FFF7FB}} \color[HTML]{000000} 14.0 & {\cellcolor[HTML]{FFF7FB}} \color[HTML]{000000} 14.8 & {\cellcolor[HTML]{FFF7FB}} \color[HTML]{000000} 14.5 \\
Harm & {\cellcolor[HTML]{FFF7FB}} \color[HTML]{000000} 16.6 & {\cellcolor[HTML]{FFF7FB}} \color[HTML]{000000} 13.1 & {\cellcolor[HTML]{FFF7FB}} \color[HTML]{000000} 15.0 & {\cellcolor[HTML]{FFF7FB}} \color[HTML]{000000} 15.0 & {\cellcolor[HTML]{FFF7FB}} \color[HTML]{000000} 12.7 & {\cellcolor[HTML]{91B5D6}} \color[HTML]{000000} 35.6 & {\cellcolor[HTML]{EDE7F2}} \color[HTML]{000000} 23.7 & {\cellcolor[HTML]{DCDAEB}} \color[HTML]{000000} 26.5 & {\cellcolor[HTML]{E2DFEE}} \color[HTML]{000000} 25.5 & {\cellcolor[HTML]{D5D5E8}} \color[HTML]{000000} 27.8 \\
Cheating & {\cellcolor[HTML]{FFF7FB}} \color[HTML]{000000} 13.3 & {\cellcolor[HTML]{FFF7FB}} \color[HTML]{000000} 16.1 & {\cellcolor[HTML]{FFF7FB}} \color[HTML]{000000} 16.2 & {\cellcolor[HTML]{FFF7FB}} \color[HTML]{000000} 14.5 & {\cellcolor[HTML]{FFF7FB}} \color[HTML]{000000} 11.0 & {\cellcolor[HTML]{EDE7F2}} \color[HTML]{000000} 23.7 & {\cellcolor[HTML]{B8C6E0}} \color[HTML]{000000} 31.4 & {\cellcolor[HTML]{B8C6E0}} \color[HTML]{000000} 31.4 & {\cellcolor[HTML]{E1DFED}} \color[HTML]{000000} 25.7 & {\cellcolor[HTML]{EAE6F1}} \color[HTML]{000000} 24.1 \\
Betrayal & {\cellcolor[HTML]{FFF7FB}} \color[HTML]{000000} 13.3 & {\cellcolor[HTML]{FFF7FB}} \color[HTML]{000000} 16.1 & {\cellcolor[HTML]{FAF2F8}} \color[HTML]{000000} 20.3 & {\cellcolor[HTML]{FFF7FB}} \color[HTML]{000000} 17.4 & {\cellcolor[HTML]{FFF7FB}} \color[HTML]{000000} 14.0 & {\cellcolor[HTML]{DCDAEB}} \color[HTML]{000000} 26.5 & {\cellcolor[HTML]{B8C6E0}} \color[HTML]{000000} 31.4 & {\cellcolor[HTML]{3D93C2}} \color[HTML]{F1F1F1} 42.6 & {\cellcolor[HTML]{ACC0DD}} \color[HTML]{000000} 32.8 & {\cellcolor[HTML]{E0DEED}} \color[HTML]{000000} 25.9 \\
Subversion & {\cellcolor[HTML]{FFF7FB}} \color[HTML]{000000} 12.3 & {\cellcolor[HTML]{FFF7FB}} \color[HTML]{000000} 16.2 & {\cellcolor[HTML]{FBF4F9}} \color[HTML]{000000} 19.9 & {\cellcolor[HTML]{FFF7FB}} \color[HTML]{000000} 19.0 & {\cellcolor[HTML]{FFF7FB}} \color[HTML]{000000} 14.8 & {\cellcolor[HTML]{E2DFEE}} \color[HTML]{000000} 25.5 & {\cellcolor[HTML]{E1DFED}} \color[HTML]{000000} 25.7 & {\cellcolor[HTML]{ACC0DD}} \color[HTML]{000000} 32.8 & {\cellcolor[HTML]{88B1D4}} \color[HTML]{000000} 36.5 & {\cellcolor[HTML]{E7E3F0}} \color[HTML]{000000} 24.6 \\
Degradation & {\cellcolor[HTML]{FFF7FB}} \color[HTML]{000000} 13.6 & {\cellcolor[HTML]{FFF7FB}} \color[HTML]{000000} 11.4 & {\cellcolor[HTML]{FFF7FB}} \color[HTML]{000000} 14.3 & {\cellcolor[HTML]{FFF7FB}} \color[HTML]{000000} 13.2 & {\cellcolor[HTML]{FFF7FB}} \color[HTML]{000000} 14.5 & {\cellcolor[HTML]{D5D5E8}} \color[HTML]{000000} 27.8 & {\cellcolor[HTML]{EAE6F1}} \color[HTML]{000000} 24.1 & {\cellcolor[HTML]{E0DEED}} \color[HTML]{000000} 25.9 & {\cellcolor[HTML]{E7E3F0}} \color[HTML]{000000} 24.6 & {\cellcolor[HTML]{A4BCDA}} \color[HTML]{000000} 33.7 \\
\end{tabular}
\caption{Moral similarity for MFD2.0 with the off-the-shelf SimCSE approach.}
\label{tab:moral-sim-sup-base-mfd2}
\end{table}

\begin{table}[!htb]
\centering
\scriptsize
\hspace{-0.7cm}
\renewcommand{\tabcolsep}{1.5pt}
\renewcommand{\arraystretch}{1.4} 
\begin{tabular}{rccccc|ccccc}
 & \rot{Care} & \rot{Fairness} & \rot{Loyalty} & \rot{Authority} & \rot{Purity} & \rot{Harm} & \rot{Cheating} & \rot{Betrayal} & \rot{Subversion} & \rot{Degradation} \\
Care & {\cellcolor[HTML]{88B1D4}} \color[HTML]{000000} 36.4 & {\cellcolor[HTML]{DBDAEB}} \color[HTML]{000000} 26.8 & {\cellcolor[HTML]{C4CBE3}} \color[HTML]{000000} 30.0 & {\cellcolor[HTML]{D2D3E7}} \color[HTML]{000000} 28.2 & {\cellcolor[HTML]{C9CEE4}} \color[HTML]{000000} 29.4 & {\cellcolor[HTML]{C5CCE3}} \color[HTML]{000000} 30.0 & {\cellcolor[HTML]{DBDAEB}} \color[HTML]{000000} 26.7 & {\cellcolor[HTML]{D2D3E7}} \color[HTML]{000000} 28.3 & {\cellcolor[HTML]{DCDAEB}} \color[HTML]{000000} 26.5 & {\cellcolor[HTML]{D3D4E7}} \color[HTML]{000000} 28.1 \\
Fairness & {\cellcolor[HTML]{DBDAEB}} \color[HTML]{000000} 26.8 & {\cellcolor[HTML]{B3C3DE}} \color[HTML]{000000} 32.1 & {\cellcolor[HTML]{D5D5E8}} \color[HTML]{000000} 27.8 & {\cellcolor[HTML]{D6D6E9}} \color[HTML]{000000} 27.7 & {\cellcolor[HTML]{D9D8EA}} \color[HTML]{000000} 27.0 & {\cellcolor[HTML]{DCDAEB}} \color[HTML]{000000} 26.5 & {\cellcolor[HTML]{D2D3E7}} \color[HTML]{000000} 28.2 & {\cellcolor[HTML]{D2D3E7}} \color[HTML]{000000} 28.2 & {\cellcolor[HTML]{D4D4E8}} \color[HTML]{000000} 27.9 & {\cellcolor[HTML]{DAD9EA}} \color[HTML]{000000} 26.8 \\
Loyalty & {\cellcolor[HTML]{C4CBE3}} \color[HTML]{000000} 30.0 & {\cellcolor[HTML]{D5D5E8}} \color[HTML]{000000} 27.8 & {\cellcolor[HTML]{75A9CF}} \color[HTML]{F1F1F1} 38.3 & {\cellcolor[HTML]{B9C6E0}} \color[HTML]{000000} 31.3 & {\cellcolor[HTML]{C5CCE3}} \color[HTML]{000000} 29.9 & {\cellcolor[HTML]{D0D1E6}} \color[HTML]{000000} 28.7 & {\cellcolor[HTML]{C8CDE4}} \color[HTML]{000000} 29.6 & {\cellcolor[HTML]{9FBAD9}} \color[HTML]{000000} 34.1 & {\cellcolor[HTML]{B3C3DE}} \color[HTML]{000000} 32.0 & {\cellcolor[HTML]{CDD0E5}} \color[HTML]{000000} 29.0 \\
Authority & {\cellcolor[HTML]{D2D3E7}} \color[HTML]{000000} 28.2 & {\cellcolor[HTML]{D6D6E9}} \color[HTML]{000000} 27.7 & {\cellcolor[HTML]{B9C6E0}} \color[HTML]{000000} 31.3 & {\cellcolor[HTML]{A2BCDA}} \color[HTML]{000000} 33.9 & {\cellcolor[HTML]{C8CDE4}} \color[HTML]{000000} 29.7 & {\cellcolor[HTML]{D2D3E7}} \color[HTML]{000000} 28.2 & {\cellcolor[HTML]{D1D2E6}} \color[HTML]{000000} 28.6 & {\cellcolor[HTML]{C0C9E2}} \color[HTML]{000000} 30.4 & {\cellcolor[HTML]{B8C6E0}} \color[HTML]{000000} 31.4 & {\cellcolor[HTML]{D3D4E7}} \color[HTML]{000000} 28.0 \\
Purity & {\cellcolor[HTML]{C9CEE4}} \color[HTML]{000000} 29.4 & {\cellcolor[HTML]{D9D8EA}} \color[HTML]{000000} 27.0 & {\cellcolor[HTML]{C5CCE3}} \color[HTML]{000000} 29.9 & {\cellcolor[HTML]{C8CDE4}} \color[HTML]{000000} 29.7 & {\cellcolor[HTML]{A5BDDB}} \color[HTML]{000000} 33.5 & {\cellcolor[HTML]{D4D4E8}} \color[HTML]{000000} 28.0 & {\cellcolor[HTML]{D8D7E9}} \color[HTML]{000000} 27.2 & {\cellcolor[HTML]{CCCFE5}} \color[HTML]{000000} 29.2 & {\cellcolor[HTML]{CED0E6}} \color[HTML]{000000} 28.8 & {\cellcolor[HTML]{CCCFE5}} \color[HTML]{000000} 29.2 \\
\hhline{~----------}
Harm & {\cellcolor[HTML]{C5CCE3}} \color[HTML]{000000} 30.0 & {\cellcolor[HTML]{DCDAEB}} \color[HTML]{000000} 26.5 & {\cellcolor[HTML]{D0D1E6}} \color[HTML]{000000} 28.7 & {\cellcolor[HTML]{D2D3E7}} \color[HTML]{000000} 28.2 & {\cellcolor[HTML]{D4D4E8}} \color[HTML]{000000} 28.0 & {\cellcolor[HTML]{9AB8D8}} \color[HTML]{000000} 34.7 & {\cellcolor[HTML]{D0D1E6}} \color[HTML]{000000} 28.7 & {\cellcolor[HTML]{BDC8E1}} \color[HTML]{000000} 30.8 & {\cellcolor[HTML]{C0C9E2}} \color[HTML]{000000} 30.4 & {\cellcolor[HTML]{C1CAE2}} \color[HTML]{000000} 30.3 \\
Cheating & {\cellcolor[HTML]{DBDAEB}} \color[HTML]{000000} 26.7 & {\cellcolor[HTML]{D2D3E7}} \color[HTML]{000000} 28.2 & {\cellcolor[HTML]{C8CDE4}} \color[HTML]{000000} 29.6 & {\cellcolor[HTML]{D1D2E6}} \color[HTML]{000000} 28.6 & {\cellcolor[HTML]{D8D7E9}} \color[HTML]{000000} 27.2 & {\cellcolor[HTML]{D0D1E6}} \color[HTML]{000000} 28.7 & {\cellcolor[HTML]{A5BDDB}} \color[HTML]{000000} 33.5 & {\cellcolor[HTML]{A4BCDA}} \color[HTML]{000000} 33.7 & {\cellcolor[HTML]{B3C3DE}} \color[HTML]{000000} 32.0 & {\cellcolor[HTML]{C6CCE3}} \color[HTML]{000000} 29.8 \\
Betrayal & {\cellcolor[HTML]{D2D3E7}} \color[HTML]{000000} 28.3 & {\cellcolor[HTML]{D2D3E7}} \color[HTML]{000000} 28.2 & {\cellcolor[HTML]{9FBAD9}} \color[HTML]{000000} 34.1 & {\cellcolor[HTML]{C0C9E2}} \color[HTML]{000000} 30.4 & {\cellcolor[HTML]{CCCFE5}} \color[HTML]{000000} 29.2 & {\cellcolor[HTML]{BDC8E1}} \color[HTML]{000000} 30.8 & {\cellcolor[HTML]{A4BCDA}} \color[HTML]{000000} 33.7 & {\cellcolor[HTML]{4897C4}} \color[HTML]{F1F1F1} 41.7 & {\cellcolor[HTML]{8BB2D4}} \color[HTML]{000000} 36.2 & {\cellcolor[HTML]{B9C6E0}} \color[HTML]{000000} 31.3 \\
Subversion & {\cellcolor[HTML]{DCDAEB}} \color[HTML]{000000} 26.5 & {\cellcolor[HTML]{D4D4E8}} \color[HTML]{000000} 27.9 & {\cellcolor[HTML]{B3C3DE}} \color[HTML]{000000} 32.0 & {\cellcolor[HTML]{B8C6E0}} \color[HTML]{000000} 31.4 & {\cellcolor[HTML]{CED0E6}} \color[HTML]{000000} 28.8 & {\cellcolor[HTML]{C0C9E2}} \color[HTML]{000000} 30.4 & {\cellcolor[HTML]{B3C3DE}} \color[HTML]{000000} 32.0 & {\cellcolor[HTML]{8BB2D4}} \color[HTML]{000000} 36.2 & {\cellcolor[HTML]{6FA7CE}} \color[HTML]{F1F1F1} 38.7 & {\cellcolor[HTML]{BBC7E0}} \color[HTML]{000000} 31.0 \\
Degradation & {\cellcolor[HTML]{D3D4E7}} \color[HTML]{000000} 28.1 & {\cellcolor[HTML]{DAD9EA}} \color[HTML]{000000} 26.8 & {\cellcolor[HTML]{CDD0E5}} \color[HTML]{000000} 29.0 & {\cellcolor[HTML]{D3D4E7}} \color[HTML]{000000} 28.0 & {\cellcolor[HTML]{CCCFE5}} \color[HTML]{000000} 29.2 & {\cellcolor[HTML]{C1CAE2}} \color[HTML]{000000} 30.3 & {\cellcolor[HTML]{C6CCE3}} \color[HTML]{000000} 29.8 & {\cellcolor[HTML]{B9C6E0}} \color[HTML]{000000} 31.3 & {\cellcolor[HTML]{BBC7E0}} \color[HTML]{000000} 31.0 & {\cellcolor[HTML]{ABBFDC}} \color[HTML]{000000} 33.0 \\
\end{tabular}
\caption{Moral similarity for MFD2.0 with the unsupervised SimCSE approach.}
\label{tab:moral-sim-mfd2-unsup}
\end{table}

\subsubsection{Classification}
\label{sec:classification-method}
\rebuttal{As suggested in the literature \cite{eger-etal-2019-pitfalls}, we test the resulting embedding spaces by adding a linear layer (i.e., a fully connected layer) with 11 output features as a classification head on top of the trained moral embedding spaces, to predict the 11 labels described in Table~\ref{tab:label_distr}.
We compare the off-the-shelf SimCSE model and the embeddings trained with unsupervised and supervised approaches to judge the effectiveness of the (un)supervised training of the moral embeddings for the classification task.
The three compared embedding spaces are not retrained---we only train the linear layer on the test set with 5-fold cross-validation and report mean and standard deviation.}
The hyperparameters used for the linear classifier are reported in Table \ref{tab:hyperparams-cls}. Default and commonly used values were chosen.

\begin{table}[!htb]
\small
\centering
\begin{tabular}{@{}ll@{}}
\toprule
\textbf{Hyperparameters} & \textbf{Options}                \\ \midrule
Max Sequence Length      & 64                     \\
Epochs                   & 10                         \\
Batch Size               & 16                      \\
Learning Rate            & 0.01 \\
Dropout                  & 0.1 \\
Loss function            & Binary Cross Entropy \\
\bottomrule
\end{tabular}
\caption{Hyperparameters used for the linear classifier.}
\label{tab:hyperparams-cls}
\end{table}

\paragraph{Results}
\label{sec:classification}

\rebuttal{We report the mean and standard deviation of the micro and macro $F_1$-scores in Table~\ref{tab:classification-6}.}

\begin{table}[!htb]
\small
\centering
\begin{tabular}{@{}lcc@{}}
\toprule
\textbf{Approach}    & \textbf{Micro $F_1$}          & \textbf{Macro $F_1$}          \\
\midrule
Supervised SimCSE   & \textbf{68.4 $\pm$ 3.1} & \textbf{56.7 $\pm$ 2.6} \\
Unsupervised SimCSE & 58.0 $\pm$ 2.9  & 36.2 $\pm$ 3.4  \\ 
\rebuttal{Off-the-shelf SimCSE}   & 59.4 $\pm$ 3.1   & 39.4 $\pm$ 3.9 \\    
\bottomrule
\end{tabular}
\caption{Classification results for the three compared approaches.}
\label{tab:classification-6}
\end{table}

\rebuttal{First, we notice that the supervised SimCSE approach clearly outperforms the off-the-shelf model and the unsupervised approach, confirming that label information is crucial to recognize a pluralist approach to morality.
Further, the reported $F_1$-scores are in line with previous experiments on the same dataset \cite{liscio-etal-2022-cross}, which we reproduce in the next section.
Second, the unsupervised approach does not improve over the off-the-shelf model despite having been exposed to the training set, showing that the necessity of labels overshadows the need for large amounts of training data for the task of pluralist moral classification.
}

\paragraph{BERT Baseline}

\rebuttal{We also add two baselines by performing multi-label classification with BERT \cite{devlin-etal-2019-bert}, which is considered state-of-the-art in the classification of the MFT taxonomy \cite{alshomary-etal-2022-moral,liscio-etal-2022-cross, Huang2022LearningWeighting,Bulla2023}. 
In the first variant (referred to as `BERT'), we first train BERT on the MFTC training set and then we continue to train it on the test set with a 5-fold cross-validation. In the second variant (referred to as `BERT (base)'), we only train BERT on the test set with a 5-fold cross-validation.}
We base the hyperparameters on the ones used by \citet{liscio-etal-2022-cross}, who performed experiments with the same corpus and model. We set the number of epochs to 10, similar to the linear classifier used in the previous experiments. 
The hyperparameters are listed in Table~\ref{tab:hyperparams-bfsc} and the results are shown in Table~\ref{tab:classification-bert}.

\begin{table}[!htb]
\small
\centering
\begin{tabular}{@{}ll@{}}
\toprule
\textbf{Hyperparameters} & \textbf{Options}              \\ \midrule
Model name               & bert-large-uncased \\
Max Sequence Length      & 64                   \\
Epochs                   & 10                      \\
Batch Size               & 16                    \\
Optimizer                & AdamW \\
Learning Rate            & \textbf{2e-5}, 5e-5 \\
Loss function            & Binary Cross Entropy \\
\bottomrule
\end{tabular}
\caption{Hyperparameters for the BERT baseline. In bold, the chosen hyperparameters.}
\label{tab:hyperparams-bfsc}
\end{table}

\begin{table}[!htb]
\small
\centering
\begin{tabular}{@{}lcc@{}}
\toprule
\textbf{Approach}    & \textbf{Micro $F_1$}          & \textbf{Macro $F_1$}          \\
\midrule
BERT    & 71.0 $\pm$ 1.5 & 62.2 $\pm$ 1.1 \\
BERT (base)   & 66.2 $\pm$ 2.4 & 55.8 $\pm$ 1.2 \\ 
\bottomrule
\end{tabular}
\caption{Classification results for the BERT baseline.}
\label{tab:classification-bert}
\end{table}

The end-to-end training of BERT offers an advantage with respect to the split training (sentence embeddings + linear classifier) of the SimCSE approaches. Further, we only choose a simple linear layer as classifier head on top of the SimCSE embeddings, yet being aware that a more complex classifier could lead to better performance. As a result, the results of the supervised SimCSE approach (Table~\ref{tab:classification-6}) are comparable to the BERT baseline in micro $F_1$-score and worse in macro $F_1$-score, showing BERT's better capacity at handling imbalanced datasets. However, the goal of the SimCSE classification evaluation is not to improve the classification performance over the BERT baselines but rather to compare the effectiveness of the different training approaches.

\paragraph{Misclassification Error Analysis} 
To further analyze the results of the five classification approaches, we inspect (1) the confusion between moral and non-moral texts and (2) the confusion between and within foundations. In Table~\ref{tab:misclass-6} we show the following four types of misclassification errors (which add up to 100\%), as previously performed for a similar classification task \cite{liscio-etal-2022-cross}.

\noindent \textbf{Error I} A tweet labeled with one or more moral values is classified as non-moral or no prediction. \\
\textbf{Error II} A tweet labeled as non-moral is classified with one or more moral values. \\
\textbf{Error III} A tweet labeled with a moral value is classified with values from other foundations. \\
\textbf{Error IV} A tweet labeled as a vice/virtue is classified as the opposite virtue/vice within that foundation. \\

\begin{table}[!htb]
\small
\centering
\begin{tabular}{@{}p{2.9cm}p{0.7cm}p{0.7cm}p{0.7cm}p{0.7cm}@{}}
\toprule
\textbf{Approach}   & \textbf{I} & \textbf{II} & \textbf{III} & \textbf{IV} \\ \midrule
Supervised SimCSE   & 50.5   & 30.6   & 17.3   & 1.60   \\
Unsupervised SimCSE & 62.9   & 24.6   & 11.3   & 1.15   \\  

\rebuttal{Off-the-shelf SimCSE}   & 62.2   & 24.8   & 11.6   & 1.40   \\
\midrule
BERT  & 28.5   & 36.9   & 30.7   & 3.86   \\
BERT (base)  & 29.3   & 38.0   & 29.8   & 2.89   \\
\bottomrule
\end{tabular}
\caption{Misclassification errors (reported as percentages over the total number of errors).}
\label{tab:misclass-6}
\end{table}

The SimCSE approaches mostly incur in Error I and Error II (i.e., distinguishing between moral and non-moral texts).
Instead, the BERT models show an approximately equal distribution of Error I, Error II, and Error III. This means that, compared to SimCSE, BERT is better at distinguishing moral vs. non-moral, but worse at predicting the correct foundation. This difference can be explained by the training procedure of BERT (which uses all labeled data points, which are mostly composed of non-moral labels) vs. supervised SimCSE (which focuses on distinguishing among the moral elements).
Finally, BERT makes more mistakes between virtue and vice within a foundation (Error IV) compared to the SimCSE approaches.

\paragraph{Training Time}

Table~\ref{tab:time-perf} displays the time needed for training the models.
Off-the-shelf SimCSE and BERT (base) are not trained on the MFTC training set, thus the first values are 0. The supervised SimCSE takes significantly less total time for the training process than BERT and than the unsupervised SimCSE (which takes longer due to the larger number of triples used during training, as described in Section~\ref{sec:method} and \ref{app:data-process}). Considering the small difference in the final $F_1$-scores (Tables~\ref{tab:classification-6} and \ref{tab:classification-bert}), there is a trade-off in using the supervised SimCSE approach. Further, the embedding space can be re-used in different applications (e.g., language classification and generation).

\begin{table}[!htb]
\small
\centering
\begin{tabular}{@{}lc@{}}
\toprule
\textbf{Approach} & \textbf{Training Time (s)} \\ 
\midrule
Supervised SimCSE    & 249 + 10                         \\
Unsupervised SimCSE  & 493 + 11                       \\ 
\rebuttal{Off-the-shelf SimCSE}    & 0 + 10        \\
\midrule
BERT          & 3521 + 327        \\
BERT (base)    & 0 + 313         \\
\bottomrule
\end{tabular}
\caption{Training time comparison. The first value shows the training time on the MFTC training set and the second value is the cross-validation on the test set.}
\label{tab:time-perf}
\end{table}

\paragraph{Per-label Classification Results}
Table~\ref{tab:detailed-simcse} and \ref{tab:detailed-bfsc} show the mean and standard deviation of $F_1$-scores for each label. Overall, a common pattern can be observed. \textit{Cheating} and \textit{harm} are the easiest vice values to classify, while \textit{fairness} and \textit{care} are the easiest virtues value to classify. On the other hand, the \textit{purity} element is always difficult to identify for all approaches, likely due to the presence of fewer examples with this label in the dataset.

\begin{table}[!htb]
\centering
\small
\begin{tabular}{@{}lcc@{}}
\toprule
 & \textbf{Sup. SimCSE}  & \textbf{Unsup. SimCSE} \\ 
 \midrule
Care        & 67.9 $\pm$ 5.2  & 56.7 $\pm$ 3.7   \\
Harm        & 57.5 $\pm$ 4.8  & 48.1 $\pm$ 6.7   \\
Fairness    & 71.4 $\pm$ 6.3  & 50.3 $\pm$8.8   \\
Cheating    & 66.0 $\pm$ 3.6  & 40.1 $\pm$ 7.7   \\
Loyalty     & 61.1 $\pm$ 6.0  & 36.7 $\pm$ 15.0   \\
Betrayal    & 51.0 $\pm$ 9.4  & 16.8 $\pm$ 3.3   \\
Authority   & 54.9 $\pm$ 10.4  & 30.2 $\pm$ 14.1   \\
Subversion  & 37.1 $\pm$ 13.1  & 16.3 $\pm$ 3.9   \\
Purity      & 46.3 $\pm$ 21.8  & 14.3 $\pm$ 10.1   \\
Degradation & 32.2 $\pm$ 12.4  & 14.6 $\pm$ 13.6   \\
Non-moral   & 78.0 $\pm$ 3.7  & 73.9 $\pm$ 3.1   \\ \bottomrule
\end{tabular}
\caption{Per-label classification mean and standard deviation for the compared SimCSE approaches.}
\label{tab:detailed-simcse}
\end{table}

\begin{table}[!htb]
\centering
\small
\begin{tabular}{@{}lcc@{}}
\toprule
& \textbf{BERT}  & \textbf{BERT (base)} \\ \midrule
Care        & 70.5 $\pm$ 4.1      & 67.0 $\pm$ 3.3    \\
Harm        & 64.7 $\pm$ 4.5      & 57.9 $\pm$ 4.3     \\
Fairness    & 70.8 $\pm$ 7.8      & 68.7 $\pm$ 6.1    \\
Cheating    & 71.2 $\pm$ 4.5      & 64.8 $\pm$ 4.9    \\
Loyalty     & 65.4 $\pm$ 4.5      & 59.9 $\pm$ 5.2    \\
Betrayal    & 55.5 $\pm$ 13.2      & 48.2 $\pm$ 9.7    \\
Authority   & 59.6 $\pm$ 7.8      & 51.5 $\pm$ 12.9    \\
Subversion  & 44.8 $\pm$ 10.2      & 39.1 $\pm$ 13.5    \\
Purity      & 50.1 $\pm$ 8.1      & 41.7 $\pm$ 10.7    \\
Degradation & 52.5 $\pm$ 14.0      & 38.4 $\pm$ 14.5    \\
Non-moral   & 80.3 $\pm$ 2.3      & 77.2 $\pm$ 3.5    \\ \bottomrule
\end{tabular}
\caption{Per-label classification mean and standard deviation for the BERT models.}
\label{tab:detailed-bfsc}
\end{table}

\paragraph{Foundations-only Results}
We additionally experimented with 6 labels, i.e., the 5 foundations (combining vices and virtues) plus the \textit{non-moral} label.
The supervised approach dataset construction slightly differs as vice and virtue from the same foundation are in this case assigned the same label. Thus, the positive instance is chosen as a data point annotated with the same foundation, and the negative instance as a data point annotated with a different foundation.

We show the results with 6 and 11 labels (as in Table~\ref{tab:classification-6}) in Table~\ref{tab:foundation-only-result}. The used hyperparameters are in Tables~\ref{tab:foundation-hyperparams-supervised} and \ref{tab:foundation-hyperparams-unsupervised}. 
We observe that the results are comparable. Since distinguishing between vice and virtue allows for a more fine-grained interpretation of morality with respect to only distinguishing among foundations, we opted for the 11-label approach.

\begin{table}[!htb]
\centering
\small
\begin{tabular}{@{}lcc@{}}
\toprule
\textbf{Approach}         & \textbf{Micro $F_1$} & \textbf{Macro $F_1$} \\ \midrule
Supervised SimCSE (6 labels)   & 68.0 & 56.7 \\
Unsupervised SimCSE (6 labels) & 57.5 & 39.4 \\ \midrule
Supervised SimCSE (11 labels)   & 68.4 & 56.7 \\
Unsupervised SimCSE (11 labels) & 58.0 & 36.2 \\ \bottomrule
\end{tabular}
\caption{Classification result with 6 and 11 labels.}
\label{tab:foundation-only-result}
\end{table}

\begin{table}[!htb]
\small
\centering
\begin{tabular}{@{}ll@{}}
\toprule
\textbf{Hyperparameters} & \textbf{Options}              \\ \midrule
Model name               & sup-simcse-bert-large-uncased \\
Max Sequence Length      & 64         \\
Epochs                   & 3                 \\
Batch Size               & 16                 \\
Learning Rate            & $5\times 10^{-5}$\\
Temperature              & 0.05           \\
Pooler                   & cls \\ \bottomrule
\end{tabular}
\caption{Hyperparameters chosen for the 6-label supervised SimCSE approach.}
\label{tab:foundation-hyperparams-supervised}
\end{table}

\begin{table}[!htb]
\small
\centering
\begin{tabular}{@{}ll@{}}
\toprule
\textbf{Hyperparameters} & \textbf{Options}                \\ \midrule
Model name               & unsup-simcse-bert-large-uncased \\
Max Sequence Length      & 64                     \\
Epochs                   & 1                      \\
Batch Size               & 16                     \\
Learning Rate            & $3 \times 10^{-5}$\\
Temperature              & 0.05                \\ 
Pooler                   & cls \\ \bottomrule
\end{tabular}
\caption{Hyperparameters chosen for the 6-label unsupervised SimCSE approach.}
\label{tab:foundation-hyperparams-unsupervised}
\end{table}
\label{sec:appendix2}

\end{document}